\documentclass[11pt]{article}

\usepackage[final]{acl}

\usepackage{times}
\usepackage{latexsym}

\usepackage[T1]{fontenc}

\usepackage[utf8]{inputenc}

\usepackage{microtype}

\usepackage{inconsolata}

\usepackage{graphicx}
\usepackage{fontawesome5} 
\usepackage{seqsplit}

\title{Skill Weaving: Efficient LLM Improvement via Modular Skillpacks}

\author{
    Zhuo Li$^{1*}$
    \quad Guodong Du$^{2*}$
    \quad Zesheng Shi$^1$
    \quad Weiyang Guo$^1$\\
    \quad \textbf{Weijun Yao}$^3$
    \quad \textbf{Yuan Zhou}$^3$
    \quad \textbf{Jiabo Zhang}$^4$
    \quad  \textbf{Jing Li}$^1$\textsuperscript{\texorpdfstring{\faIcon[regular]{envelope}}{}} 
    \\$^{1}$Harbin Institute of Technology, Shenzhen, China 
    \\$^{2}$The Hong Kong Polytechnic University   
    \\$^{3}$Huawei Technologies Co., Ltd. \quad
     $^{4}$Shanghai Jiaotong University \\
    \texttt{zuoer190191@mail.ustc.edu.cn} \quad \texttt{jingli.phd@hotmail.com}  
}

\usepackage[utf8]{inputenc} 
\usepackage[T1]{fontenc}    
\usepackage{hyperref}       
\usepackage{url}            
\usepackage{booktabs}       
\usepackage{amsfonts}       
\usepackage{nicefrac}       
\usepackage{microtype}      
\usepackage{xcolor}         

\usepackage[textsize=tiny]{todonotes}

\usepackage{wrapfig}
\usepackage{amsmath}
\usepackage{amssymb}
\usepackage{mathtools}
\usepackage{amsthm}

\usepackage[table]{xcolor}

\usepackage{algorithm}
\usepackage{algorithmic}

\usepackage{amsmath}

\usepackage{multirow}
\usepackage{pifont}
\usepackage{graphicx} 
\usepackage{xspace}

\newcommand{\pub}[1]{{\color{gray}{\tiny{[{#1}]\!}}}}

\definecolor{mygray}{gray}{.9}
\definecolor{ggray}{RGB}{127,127,127}
\definecolor{redb}{RGB}{217,148,143}
\definecolor{myyellow}{RGB}{190,144,0}
\definecolor{mygreen}{RGB}{80,100,40}
\definecolor{tabhighlight}{HTML}{e5e5e5}
\definecolor{cyan}{HTML}{00B075}

\usepackage{enumitem}

\newcommand{\ie}{\emph{i.e.,}\xspace}
\newcommand{\eg}{\emph{e.g.,}\xspace}

\newcommand{\ourapproach}{\texttt{SkillWeave}\xspace}
\newcommand{\quantapproach}{\texttt{SKillZip}\xspace}

\usepackage[utf8]{inputenc} 
\usepackage{url}            
\usepackage{booktabs}       
\usepackage{amsfonts}       
\usepackage{nicefrac}       
\usepackage{microtype}      
\usepackage{tabularx}       
\usepackage{bm}
\usepackage{enumitem}

\usepackage{adjustbox}
\usepackage{multirow}
\usepackage{pifont}
\usepackage{graphicx} 
\usepackage{xspace}
\usepackage{amsmath}
\usepackage{amssymb}
\usepackage{mathtools}
\usepackage{amsthm}
\usepackage{pifont}
\usepackage{caption}
\usepackage{makecell}              
\usepackage{nicematrix}
\usepackage{listings}
\usepackage{xtab}
\usepackage{arydshln}
\usepackage{colortbl} 
\usepackage[table, dvipsnames]{xcolor}  
\usepackage{array}                 
\usepackage{color} 
\usepackage{nicematrix}             

\definecolor{dark1}{gray}{0.85}

\usepackage[normalem]{ulem}

\usepackage[utf8]{inputenc} 
\usepackage{titletoc}

\definecolor{revisioncolor}{RGB}{0,0,255}

\definecolor{reda}{RGB}{192,0,0}

\begin{document}
\maketitle
\begin{abstract}
Large language models increasingly require specialization across diverse domains, yet existing approaches struggle to balance multi-domain capacities with strict memory and inference constraints.
In this work, we introduce SkillWeave, a modular improvement framework that enables LLMs to specialize under fixed memory budgets. SkillWeave partitions full capabilities of a general-purpose model into skillpacks—lightweight, domain-specific delta modules—that reorganize and refine the model’s internal knowledge. For efficient deployment, SkillWeave integrates SkillZip to compress skillpacks into compact and inference-ready format, enabling strong multi-domain performance with low-latency execution. On multi-task and agentic benchmarks, a 9B SkillWeave model outperforms several baselines and even surpasses a 32B monolithic LLM, while achieving up to 4× speedup.

\let\thefootnote\relax\footnotetext{\faIcon[regular]{envelope}~Corresponding author. $*$~Equal contribution.}	
\let\thefootnote\relax\footnotetext{Code will be available at \href{https://github.com/lizhuolz/SkillWeave}{code}}

\end{abstract}

\section{Introduction}

Large Language Models (LLMs) pretrained on diverse corpora demonstrate strong general-purpose capabilities and can be further adapted to downstream tasks~\citep{data0,data1,add5jailbreak,add3safety,add4backdoors}. However, as application scenarios grow increasingly diverse, a single monolithic model struggles to simultaneously achieve high performance across heterogeneous domains~\citep{self0,self1}, which in practice forces practitioners to maintain multiple specialized models through repeated post-training. A straightforward alternative is to employ larger, more general LLMs with stronger capacity across a broad range of tasks; yet this approach incurs substantial costs in memory footprint and inference latency, making it impractical for many real-world deployments~\citep{delta-come}. These challenges motivate a central question: \textbf{\emph{how can we enable LLMs to sustain multi-domain performance under fixed memory and latency budgets, with minimal resources?}}

To address these challenges, we propose \ourapproach, a modular framework for efficiently improving LLMs under fixed memory and inference budgets. At a high level, \ourapproach partitions the model’s full capability space into a set of domain-specialized \emph{skillpacks} that are independently trained, compactly stored, and flexibly deployed. \uline{First}, the base model is independently fine-tuned for each domain on self-generated data, producing a collection of domain-specific skill representations that capture improvements in specialized capabilities.  \uline{Second}, the fully fine-tuned skill parameters are compressed into a lightweight and inference-efficient format via SkillZip, resulting in a general-purpose shared backbone together with multiple specialized skillpacks suitable for compact storage and deployment. \uline{Finally}, at inference time, a shared backbone remains active while a single skillpack is dynamically selected, serving multiple specialized capabilities within a single system. 

\textbf{The central innovation of \ourapproach} is to decompose the model’s monolithic capabilities into a diverse set of domain-specific skillpacks, then compact them back into deployable form. Our design addresses two fundamental limitations overlooked in prior work. \uline{\textbf{First.}} Full-parameter fine-tuning enables high capacity but incurs prohibitive storage and inference overhead, while PEFT~\citep{peft} methods (\eg LoRA~\citep{lora}) is lightweight but often underperform and fail to retain sufficient ability of specific domain. \uline{In contrast}, \ourapproach resolves this trade-off by first training each skillpack with the full parameter budget to absorb fine-grained ability, followed by compact compression for deployment. \uline{\textbf{Second.}} Monolithic LLMs entangle all domains within a single shared parameter space, while such space sharing imposes a bottleneck due to task interference, leading to suboptimal performance and catastrophic forgetting~\citep{ties,pcb,add2graftllm,add1NeuralParameter}. \uline{In contrast}, \ourapproach decompose the LLM’s overall capacity into skillpacks and isolates training by domain, producing independent skillpacks that preserve refined performance on each domain while avoiding cross-task interference.

While domain skillpacks enable fine-grained specialization, naïvely retaining multiple full-parameter skillpacks is impractical due to storage and inference costs. To close this gap, \textbf{we propose \quantapproach, a compression strategy tailored for lightweight and inference-efficient deployment}. By first merging shared knowledge into a common backbone, \quantapproach disentangles task-specific skillpacks and reinforces core capabilities. At its core, it employs a fully quantized design that compresses both weights and activations, eliminating the need for runtime decompression or costly dequantization. This design significantly reduces inference latency, contrasts with prior delta-compression methods~\citep{asvd,delta-come} that primarily reduce storage size. In this way, SkillZip preserves the benefits of full-parameter specialization while achieving low-latency, resource-efficient inference, comparable to the original backbone model, completing the SkillWeaving pipeline from capability acquisition to deployment.

In summary, this paper makes \textbf{three significant contributions:}


\begin{itemize}
\setlength{\itemindent}{-5mm}

    \item  
    We propose \ourapproach, a modular improvement framework that enables an LLM to enhance itself under fixed memory and inference budgets. By decomposing model capabilities into compact skillpacks, \ourapproach reorganize and refine the model’s internal knowledge structure, achieving scalable and interpretable self-improvement. 

    \item We introduce \quantapproach, a fully quantized delta compression strategy tailored for modular deployment. Unlike prior approaches focused on weight storage, \quantapproach jointly quantizes weights and activations, eliminating runtime decompression and delivering low-latency inference with hardware-aware optimization.


    \item  We validate our method on multi-task and agentic benchmarks, where a 9B SkillWeave model outperforms task-specific models and a 32B monolithic model, while achieving 4× faster inference speedup and superior fidelity compared to existing delta-compression baselines.

    
\end{itemize}
\begin{figure*}[t]
    \centering \includegraphics[width=0.99\linewidth, trim=0 2 2 0, clip]{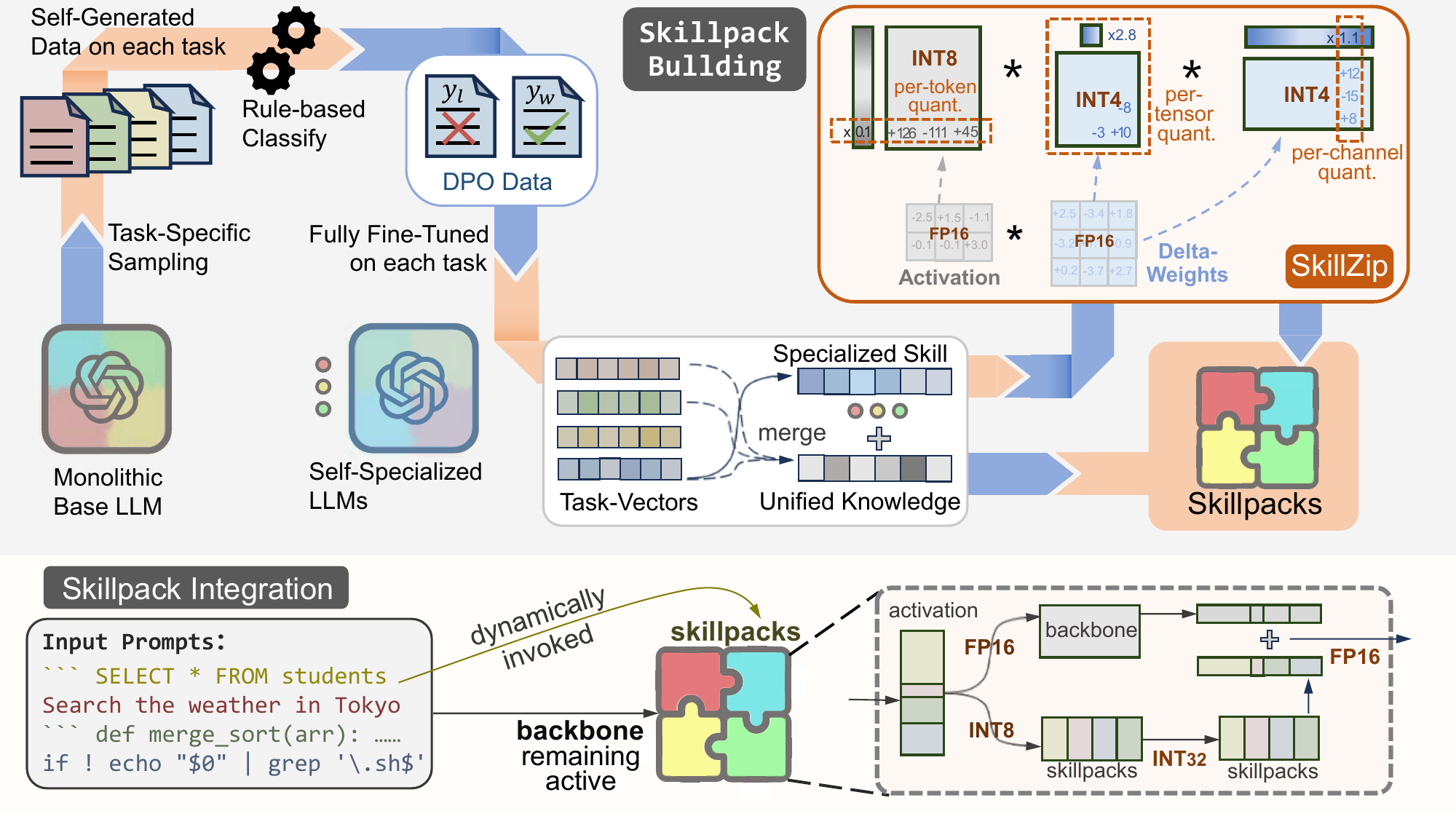}
    \vspace{-5pt}
    \caption{Overview of the \ourapproach framework. The \textbf{top} section illustrates the full pipeline, consisting of two stages: (1) decomposing a monolithic language model into task-specific skill vectors via preference-based training, and (2) compressing each skillpack into an inference-friendly form using structured quantization. The \textbf{bottom} section demonstrates the integration and inference process under an agent-style serving scenario. 
    As a whole, the figure shows how \ourapproach reorganizes and refines the model’s internal knowledge structure to balance performance gains and parameter efficiency.
    \label{fig:main figure}
    }
    \vspace{-15pt}
\end{figure*}
\section{Related Work}
\label{sec:related Work}
\subsection{Self Improvement via Synthetic Data}
 A line of recent work explores improving LLMs using self-generated synthetic data, aiming to reduce reliance on human annotations or external teachers. Self-Specialization~\citep{self-special} fine-tunes models on their own generations to induce task-specific behaviors, while Self-MoE~\citep{self-moe} extends this idea by routing inputs to independently self-specialized LoRA experts. 
Another group of methods prompts LLM to self-annotate synthetic data by itself. 
Representative examples include Self-Rewarding~\citep{self-rewarding}, Meta-Rewarding~\citep{meta-rewarding}, Self-Align~\citep{Self-Align} and RLAIF~\citep{RLAIF}, which adapt DPO~\citep{dpo} or RLHF pipeline with synthetic feedback or rule-based guidance.

\subsection{Task Vector Merging and Compression}

Task Arithmetic ~\citep{ta} first  introduced the concept of task vectors - defined as the difference between a fine-tuned model and its base. Subsequent works (\eg Ties-Merging ~\citep{ties}, DARE ~\citep{dare} and PCB-Merging ~\citep{pcb}) applied it to merging large language models, while more recent efforts further tackle task interference and dynamic deployment from neuronal and input-adaptive perspectives~\citep{fang2025disentangling,du2026didi}.

Meanwhile, recent works have explored delta-compression to reduce the overhead of storing and serving multiple task vectors. BitDelta~\citep{bitdelta} introduces 1-bit quantization for delta weights with scaling factors, significantly reducing storage. 
SVD-based methods, such as SVD-LLM~\citep{svd-llm}, ASVD~\citep{asvd} and Twin-Merging~\citep{twin}, leverage low-rank decomposition to compress task-specific deltas~\citep{svd2,taskSVD}. 
Yet, these methods are reported to yield suboptimal accuracy. 
DeltaCome~\citep{delta-come}, GPT-Zip~\citep{gptzip} and D-QRELO~\citep{li2026d} further combine quantization with sparsification or residual low-rank approximation, but sparsity alone offers limited benefit under current hardware constraints at inference.

\section{Methodology}
\paragraph{Problem Setting.} In this section, we introduce \ourapproach, a modular self-improvement framework that decomposes monolithic language models into compact, self-specialized \emph{skillpacks}. Each skillpack satisfy three criteria: (i) it accurately captures a distinct capability learned from self-generated data, (ii) it preserves general competence by avoiding interference across domains, and (iii) it can be deployed efficiently alongside other skillpacks on modern inference engines. 

\paragraph{Overview of Pipeline.} \ourapproach realizes this objective through a structured three-stage pipeline: 
(1) In Section~\ref{subsec:skillpack_building}, we present Skillpack Building (Self-Specialization), where the base model is independently specialized for each target capability to extract domain-specific improvements. 
(2) In Section~\ref{subsec:skillpack_compression}, we introduce Skillpack Compression (Full-tuning–then-zip), which unifies shared knowledge and compresses the resulting specialized parameters into compact, inference-efficient skillpacks.
(3) In Section~\ref{subsec:skillpack_integration}, we describe Skillpack Integration (Modular Deployment), where a shared backbone is combined with a selected skillpack at inference time to enable scalable multi-capability serving.

\subsection{Skillpack Building}
\label{subsec:skillpack_building}


\paragraph{Task Decomposition.}
We assume access to a base language model $\theta_0$ and a small seed instruction dataset $\mathcal{D}_{\text{seed}}$ that covers the model’s core capabilities. 
To enable domain-wise self-specialization, we partition $\mathcal{D}_{\text{seed}}$ into $K$ disjoint subsets $\{\mathcal{D}_1, \dots, \mathcal{D}_K\}$.
Each subset is associated with a specific task $\mathcal{T}_k$ such as dialogue, reasoning, or web search, allowing domain signals to be isolated during training.


\paragraph{Self-Generation.}  For each task $\mathcal{T}_k$, we prompt the base LLM $\mathcal{M}_{\theta_0}$ with instructions from $\mathcal{D}_k$ to generate candidate responses.  This yields a self-generated dataset $\mathcal{D}_k^{\text{gen}} = \{(x_i, y_i^{\text{gen}})\}$ containing model completions for the task-specific instructions. The generated responses may vary significantly in quality, containing both valuable and potentially harmful outputs.

\paragraph{Rule-Based Classification.}  Unlike prior self-improving methods~\citep{self-special,self-moe} that treat all synthetic data as equally reliable, we apply a lightweight rule-based filter that separates $\mathcal{D}_k^{\text{gen}}$ into two preference-ranked subsets:
$\mathcal{D}_k^{+} = \text{Helpful samples}, \quad \mathcal{D}_k^{-} = \text{Harmful samples}.$ 
The rules capture coarse-grained failure patterns at the task level, such as instruction misalignment in dialogue, logical contradictions in reasoning, or test failures in coding, aiming to prevent clearly erroneous behaviors from contaminating the self-alignment process. 
\vspace{-2pt}
\paragraph{Preference Optimization.} Given the preference-labeled dataset $(\mathcal{D}_k^{+}, \mathcal{D}_k^{-})$, we fine-tune the base LLM using online Direct Preference Optimization (DPO)~\citep{dpo}. 
DPO optimizes a contrastive objective that encourages the model to prefer helpful over harmful outputs:
\vspace{-2pt}
\begin{align}
&\mathcal{L}_{\text{DPO}}(\pi_\theta; \pi_{\text{ref}}) = -\mathbb{E}_{(x, y_w, y_l) \sim \mathcal{D}} \\
&[\log \sigma(\beta \log \frac{\pi_\theta(y_w \mid x)}{\pi_{\text{ref}}(y_w \mid x)} 
 - \beta\log\frac{\pi_\theta(y_l \mid x)}{\pi_{\text{ref}}(y_l \mid x)})].  \notag
\end{align} 
This yields a new model $\mathcal{M}_k = \mathcal{M}_0 + \Delta_k$, where $\Delta_k$ captures the task-specific improvement induced by preference-guided self-training. We treat $\Delta_k$ as a \emph{proto-skillpack}, which will later be compressed and integrated.


\subsection{Skillpacks Compression}
\label{subsec:skillpack_compression}

\paragraph{From Task Vectors to Skillpacks.} Naively retaining full-parameter task deltas incurs prohibitive costs in storage and inference. To make modular specialization feasible in practice, we introduce \textbf{\quantapproach}, an inference-efficient compression strategy based on \emph{full quantization for delta computation}, that converts task-specific deltas into compact, deployable skillpacks.

\paragraph{Model Merging for Shared Knowledge.}
Before compression, we extract shared cross-task knowledge across task-specific deltas to  isolate task-specific skills. Concretely, we compute a shared component through model-merging: $\Delta_{\text{shared}} = \text{Merge}([\Delta_1, \dots, \Delta_k])$, integrate them back into the model’s backbone and subtract it from each delta:
$\Delta_i \leftarrow \Delta_i  - \Delta_{\text{shared}}$.
This enhances the backbone with generalized knowledge while making individual deltas sparser and more task-specific, facilitating subsequent compression.

\paragraph{Full Quantization for Delta-Compression.}

Existing delta-compression methods quantize only the weight parameters, and require dequantization back to floating-point (e.g., FP16) during inference. In contrast, \quantapproach applies \emph{full-quantization} to delta compression, quantizing both the delta weights and their corresponding activation inputs. 
This design allows direct computation in low-bit integer formats (e.g., INT8 or INT4), achieving significantly higher inference.



Formally, consider a linear delta weight matrix $W$ and input activations $X$. We seek a $k$-bit static quantizer $\mathrm{Quant}_k(\cdot, q)$ and a low-rank factorization $W \approx AB$, where $A \in \mathbb{R}^{C_i \times R}$ and $B \in \mathbb{R}^{R \times C_o}$. The quantized low-bit representations $\hat{X}, \hat{A}, \hat{B}$ are optimized to minimize the end-to-end reconstruction error:
\vspace{-2pt}
\begin{align}
& \mathop{\min}_{\hat{X}, \hat{A}, \hat{B}} {\| XW - \text{Dequant}(\hat{X}\hat{A}\hat{B}) \|}, \notag \\
&\hat{X}, \hat{A}, \hat{B} =\! \text{Quant}_k(X, A, B, q)  
\end{align} 
\vspace{-10pt}

Due to practical constraints of hardware-accelerated GEMM~\citep{GEMM} kernels, FP16 scaling factors can only be efficiently applied along the outer dimensions of INT8 matrix multiplication. \footnote{before and after INT8 matmul, but not between intermediate INT8 multiplications (\ie $\hat{A} \cdot \text{diag}(\vec{s}_A) \cdot \text{diag}(\vec{s}_B) \cdot \hat{B} $)}
As a result, we adopt per-token or per-tensor quantization for $\hat{X}$, per-tensor quantization for $\hat{A}$, and per-tensor or per-channel quantization for $\hat{B}$. 
Let $\vec{s}_X$ and $\vec{s}_B$ denote the corresponding per-token/per-channel scaling vectors, and $s_A$ a scalar per-tensor scale for $\hat{A}$. The reconstructed output is computed as:
\vspace{-5pt}
\begin{align}
Y &= XAB \ \approx\ \text{diag}(\vec{s}_X)\!\cdot\!\hat{X}\!\cdot\!s_A\!\cdot\!\hat{A}\!\cdot\!\hat{B}\cdot\!\text{diag}(\vec{s}_B)  \ \notag\\
&=  \text{diag}(s_A\!\cdot\!\vec{s}_X) \cdot \hat{X} \hat{A} \hat{B} \cdot \text{diag}(\vec{s}_B).  
\end{align}


This formulation preserves the functional effect of the original delta while enabling low-bit execution of the compressed skillpack. 


\begin{figure}[h]
    \vspace{-10pt}
    \centering \includegraphics[width=1.0\linewidth, trim=120 25 125 40, clip]{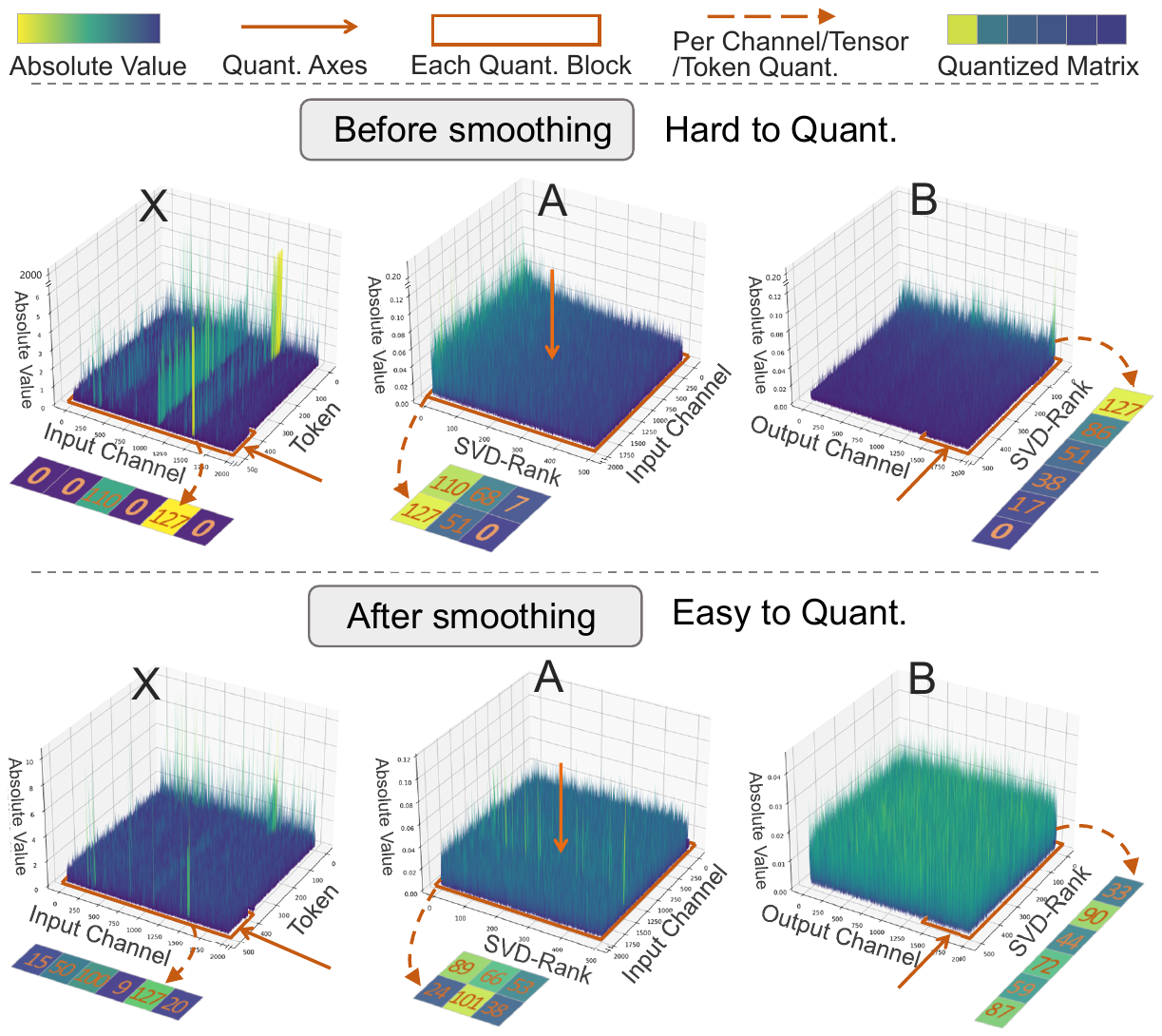}
    \vspace{-10pt}
    \caption{
    The activation $X$ and the low-rank weight matrices $A$ and $B$ are initially hard to quantize due to outliers misaligned with quantization axes. By normalizing channel scales of $X$ and spreading singular value energy across $A$ and $B$, the resulting matrices become significantly more quantization-friendly.
    \label{fig:quant}
    }
\vspace{-10pt}
\end{figure}
\paragraph{Double Smoothing for Quantization Fidelity.}
In the quantization field, accuracy degradation often arises due to outliers — a small number of activation channels with magnitudes 100× larger than the rest. 
Such outliers inflate the dynamic range of quantization and dominate reconstruction error. SkillZip addresses this challenge through a double smoothing strategy. 


First, we apply \textbf{channel-wise smoothing} to reduce outliers in the activation channels. 
As shown in Fig.~\ref{fig:quant}, activation outlier channels display \emph{task-specific patterns} that remain remarkably stable across inputs within the same domain. Inspired by SmoothQuant~\citep{smoothquant}, We learn a task-specific scale vector $s \in \mathbb{R}^{C_i}$ that rebalances the activation and weight magnitudes: 
$Y = (X \cdot \text{diag}(s)^{-1}) \cdot (\text{diag}(s) \cdot W) = X_{\text{smooth}} \cdot W_{\text{smooth}}.$
This transformation shifts some of the quantization burden from activation to weight, enabling both to be quantized more uniformly.

Next, we apply truncated singular value decomposition (SVD) to the smoothed weight matrix $W_{\text{smooth}}$:
$W_{\text{smooth}} \approx U_R \Sigma_R V_R^T, \quad A = U_R \Sigma_R^{1/2}, \quad B = \Sigma_R^{1/2} V_R^T.$
However, the concentration of energy in top singular vectors (\ie the first few columns in $A$ and rows in $B$) leads to new outliers misaligned with quantization axes (per-tensor/per-channel).
To alleviate this, we apply a second-stage \textbf{rank-wise smoothing}: a appropriate orthogonal rotation matrix $Q$ to disperses the energy evenly across dimensions:
$A_{\text{rot}} = A Q, \quad B_{\text{rot}} = Q^T B.$

Together, as demonstrated in Fig.~\ref{fig:quant}, these two smoothing steps significantly reduce quantization distortion without loss of information.

\subsection{Skillpack Integration}
\label{subsec:skillpack_integration}
At inference, our architecture maintains a unified yet modular computation flow. The shared backbone of the model remains constantly active, ensuring a stable foundation across tasks. In parallel, task-specific skillpacks are dynamically invoked based on the task type inferred from each input.\footnote{Routing implementation is detailed in Appendix~\ref{appendix:routing}}

Concretely, each Transformer block is equipped with shared weights $W$ and multiple task-specific low-rank matrices $\hat{A}_i, \hat{B}_i$. Given input activations $X$, the shared computation $XW$ proceeds as usual—either in FP16 or quantized precision. Meanwhile, tokens are grouped by task type into $\{X_1, X_2, \dots, X_K\}$, and each group is processed with its corresponding skillpack: 
\begin{align} 
\text{Output}_i = X_i W + \hat{X}_i \hat{A}_i \hat{B}_i.
\end{align} 

\section{Experimental Setup}
\subsection{Baseline Methods}

We compare \ourapproach against a comprehensive set of baselines spanning both \textbf{single-skill} and \textbf{multi-skill} self-improvement settings. 

\paragraph{Single-Skill Baselines.}
In the single-skill setting, we evaluate each skillpack independently on its corresponding domain. We consider two categories of baselines: 
\begin{itemize}[leftmargin=1.5em]
\setlength{\itemindent}{-3mm}
    \item \textbf{Alternative \emph{parameter formats} baselines.}
    \setlength{\leftmargin}{-6pt}
    1) \uline{PEFT}: Replace our full-parameter-tuning-then-SkillZip pipeline with LoRA fine-tuning. 2) \uline{Full-Parameter Finetuning}. 3)\uline{Delta-Compression}: State-of-the-art delta-compression methods,  such as BitDelta~\citep{bitdelta}, SVD-based methods~\citep{svd-llm,asvd}, and DeltaCome~\citep{delta-come}.
    

    \item \textbf{Alternative \emph{self-improvement algorithms}.} 4) \uline{Self-Specialization}~\citep{self-special}. 5) \uline{Self-Rewarding} based methods: Self-rewarding~\citep{self-rewarding}, Self-Align~\citep{Self-Align}.
\end{itemize}

\paragraph{Multi-Skill Baselines.}
The multi-skill setting is the primary application scenario for \ourapproach, where all skillpacks are simultaneously evaluated across diverse tasks. We consider a broad range of competitive baselines:

6) \uline{Open-Source LLMs}. 
7) \uline{Multi-Teacher Distillation}: FuseLLM~\citep{fusechat2} and FuseChat3.0~\citep{fusechat3}, as a upper bound.
8) \uline{Routing-Based Methods}: Self-MoE~\citep{self-moe}, LoRA-MoE~\citep{LoraMoE1} and Twin-Merging~\citep{twin}, the most relevant methods to ours. 
9) \uline{Multi-Task Learning}: jointly or sequentially fine-tuned with same training data. 
10) \uline{Model Merging and Model Grafting}: Task Arithmetic~\citep{ta}, Ties-Merging~\citep{ties} and others.

Although baselines (8)–(10) were not originally designed for self-improvement, we adapt them accordingly to ensure fair comparison. Each relies only on self-generated data and self-driven optimization, consistent with our setting. 





\subsection{Evaluation Scenarios and Benchmarks}

We design experiments across two key evaluation scenarios: (1) a general-purpose \textbf{multi-capacity} setting, and (2) a practical \textbf{LLM-as-Agent} deployment setting. These reflect both academic and real-world uses of modular, skill-based language models.

\paragraph{General Capabilities Evaluation.}
In this setting, we define four core tasks representative of general LLM capabilities: \emph{dialogue}, \emph{reasoning}, \emph{math}, and \emph{coding}. Each domain is evaluated using multiple established benchmarks. 
These tasks cover a broad spectrum of instruction-following abilities, ensuring that our evaluation is both comprehensive and robust.

\paragraph{LLM-as-Agent Evaluation.}
To evaluate this setting, we adopt \textbf{AgentBench}~\citep{agentbench}, and focus on 5 diverse and representative domains of AgentBench: \emph{Database (DB)}, \emph{Operating System (OS)}, \emph{Knowledge Graph (KG)}, \emph{Web Shopping (WS)}, and \emph{Web Browsing (WB)}. These tasks are tailored to assess LLMs acting as autonomous agents across a wide range of environments.

\paragraph{Model Instantiations.}
For most multi-task experiments, we adopt \texttt{LLaMA-3.1-8B-Instruct} as our base model, and include results on \texttt{Llama-3.2-1B-Instruct} to demonstrate the scalability and efficiency for smaller models. For the LLM-as-an-Agent experiments, we use \texttt{Qwen-2.5-7B-Instruct} as the main backbone.

\begin{table*}[!t]
\caption{
Overall performance of \ourapproach and multi-skill baselines in the general capability setting using Llama-3.1-8B-Instruct as the backbone. The best results for baselines are shown in \textbf{bold}, our results are highlighted with a pink background, and performance gaps between the best and \ourapproach are marked in \textbf{green}.
\label{tab:main_result}
}
\centering
\vspace{-2.5pt}
\resizebox{0.999\linewidth}{!}{
    \begin{NiceTabular}{@{}l|l|cc|cc|cc|cc@{}}
    \toprule
        \rowcolor{gray!20}
        \multirow{2}{*}{\textbf{Method}} & \multirow{2}{*}{\textbf{\#Params}} & \multicolumn{2}{c}{\textbf{Mathematics}} & \multicolumn{2}{c}{\textbf{Coding}} & \multicolumn{2}{c}{\textbf{Dialogue}} & \multicolumn{2}{c}{\textbf{Reasoning}}\\ 
        \cmidrule(lr){3-4} \cmidrule(lr){5-6} \cmidrule(lr){7-8} \cmidrule(lr){9-10}
        \rowcolor{gray!20}
        ~ & ~ & \textbf{GSM8k} & \textbf{MATH} & \makebox[0.08\textwidth][l]{\textbf{HumanEval}} & \makebox[0.08\textwidth][r]{\textbf{MBPP}} & \makebox[0.09\textwidth][l]{\textbf{AlpacaEval2}} & \makebox[0.08\textwidth][r]{\textbf{IFEval}} & \textbf{BBH} & \textbf{ARC-C} \\
        \midrule
        \multicolumn{10}{c}{\textbf{Open-Source LLMs}} \\ \midrule
        Llama3.1-8B-Instruct & 8B & 84.5	& 51.9	& 69.5	& 75.4	& 28.3	& 75.9	& 65.8	& 82.4 \\ 
        Qwen1.5-72B-Chat & 72B & 82.7	& 42.5	& 71.3	& 71.9  & 40.6 & 77.1 & 68.3 & 75.8 \\
        Gemma2-27B-it	& 27B & 90.4	& 54.4	& 78.7	& 81.0 & 58.9 & 77.1 & 74.9 & 77.4 \\
        Qwen2.5-14B		& 14B & 90.2	& 55.6	& 56.7	& 76.7 & 36.4 & 59.9 & 73.0 & 78.3 \\
        Qwen2-57BA14B-it	& 52B & 85.3	& 49.1	& 79.9	& 70.9 & 46.4 & 78.0 & 84.5 & 86.6 \\ \midrule
        \multicolumn{10}{c}{\textbf{Model Merging (with same fine-tuned LLMs)}} \\ \midrule
        Task Arithmetic\pub{ICLR23} & 8B  & 86.4 & 52.4	& 70.6	& 75.9	& 29.2	& 76.2	& 68.9	& 83.6 \\
        Ties-Merging\pub{NeurIPS23} & 8B  & 87.2 & 56.3	& 71.1	& 76.1	& 33.9	& 76.7	& 70.1	& 84.0 \\
        PCB-Merging\pub{NeurIPS24} & 8B   & 87.7 & 56.7	& 71.3	& 76.2	& 34.8	& 76.9	& 70.9	& 84.2 \\
        PCB-Merge+DARE\pub{ICML24} & 8B  & 88.9	& 57 & 71.5	& 76.2	& 35.0	& 77.2	& 71.2	& 84.6 \\
        \midrule
        \multicolumn{10}{c}{\textbf{Routing based LLM and Model Grafting }} \\ \midrule
        Self-MoE\pub{ACL25} & 9B  & 87	& 49.5	& 70.6	& 71.2	& 38.8	& 77.9	& 67.8	& 84.6 \\
        Routed LoRA r512 &14.1B  & 86.4	& 51.4	& 72.5	& 75.9	& 41 & 76.7	& 68.8 & 85.7 \\
        Routed LoRA r1024 & 21B  & 87.9	& 54.6	& 73.2	& 76.4	& 47.1	& 77.6	& 71.2	& 86.5 \\
        TALL-Mask\pub{ICML24} & 16.7B & 90.1 & 58.9	& 72.8	& \underline{76.2}	& 48.6	& 77.8	& 74.6	& 86.4 \\
        EMR-Merging\pub{NeurIPS24} & 16.7B  & \textbf{90.8} & \underline{59.4}	& \underline{73.5}	& 75.7	& 48.9	& 77.9	& \textbf{75.8}	& \underline{86.9}  \\
        Twin-Merging r512\pub{NeurIPS24} & 14.1B  & 87.6	& 59.3	& 73.2	& 75.5	& 46.6	& 77.7	& 71.5	& 86.7 \\
        Twin-Merging r1024 & 21B  & \underline{89.3}	& \textbf{60.4}	& \textbf{73.8}	& \textbf{76.5}	& \underline{49.6}	& \underline{78.4}	& \underline{75.3}	& \textbf{87.4}  \\
        \midrule
        \multicolumn{10}{c}{\textbf{Multi-teacher Distillation}} \\ \midrule
        FuseLLM\pub{ICLR24} & 8B  & 85.6 & 52.9	& 73.2	& 70.8	& 32.5	& 76.9	& 66.6	& 83.5 \\
        FuseChat3.0\pub{ICLR25} & 8B  & 88	& 57.2	& 71.3	& 71.8	& \textbf{64.2}	& \textbf{80.2}	& 69.4	& 82.2 \\
        \midrule
        \multicolumn{10}{c}{\textbf{Self-Rewarding}} \\ \midrule
        Self-Rewarding\pub{ICLR23} & 8B  & 84.3	& 49.7	& 68.8	& 74.9	& 46.4	& 77.9	& 66.4	& 83.6 \\
        Self-Align\pub{ACL24} & 8B  & 85.5	& 47.4	& 69.2	& 73.3	& 47.4	& 78.2	& 65.5	& 83 \\
        \midrule
        \multicolumn{10}{c}{\textbf{Multi-Task Learning (with same training data)}} \\ \midrule
        Jointly MTL & 8B   & 87.5	& 53.7	& 70.7	& 75.8	& 37.9	& 77.8	& 67.3	& 83.2\\
        Sequentially MTL & 8B  & 86.7	& 52.1	& 69.4	& 75.2	& 39.2	& 78.6	& 76.3	& 88.2\\
        \midrule
        \multicolumn{10}{c}{\textbf{Our Approach + Optional Replacement}} \\
        \midrule
        \rowcolor{pink!20}
        \textbf{SKillWeave (Ours)} & 10B  & \makebox[0.08\textwidth][c]{\textbf{91.0\textcolor{cyan}{(+0.7)}}} & \textbf{62.5\textcolor{cyan}{(+3.1)}} & \makebox[0.08\textwidth][c]{\textbf{75.0\textcolor{cyan}{(+1.5)}}} & \textbf{77.8\textcolor{cyan}{(+1.6)}} & \makebox[0.08\textwidth][c]{\textbf{52.8\textcolor{cyan}{(+3.2)}}} & \textbf{79.1\textcolor{cyan}{(+0.7)}} & \makebox[0.08\textwidth][c]{\textbf{76.2\textcolor{cyan}{(+0.9)}}} &  \textbf{88.6\textcolor{cyan}{(+1.2)}} \\
        $\rightarrow$Self-Rewarding\pub{ICLR23} & 10B  & 89.2	& 54.4	& 70.3	& 76.3	& 50.8	& 78.8	& 72.4	& 84.3 \\
        $\rightarrow$Self-Specialize\pub{ACL24} & 10B  & 87.3	& 51	& 70.7	& 72.3	& 35.6	& 76.6	& 68.3	& 85.7 \\
        $\rightarrow$PEFT & 10B  & 86.8	& 49.3	& 73.5	& 74.2	& 47.9	& 78.1	& 68.5	& 86.9 \\
        $\rightarrow$ASVD & 10B  & 89.7	& 60.5	& 73.7	& 76.9	& 47.0	& 78.3	& 74.3	& 87.2 \\
        $\rightarrow$Delta-Come\pub{NeurIPS24} & 10B  & 90.7	& 62.4	& 74.9	& 78	& 52.7	& 79	& 76.3	& 88.4 \\
        \midrule

    \end{NiceTabular}
}
\vspace{-0.4cm}
\end{table*}

\section{Main Results and Analysis}

\subsection{General Capability Results}

Tab.~\ref{tab:main_result} reports general multi-capability evaluation results across five categories of baselines. Under comparable model sizes, \ourapproach consistently achieves the strongest overall performance, ranking first in four out of five evaluated domains. 

Notably, an 8B backbone enhanced by \ourapproach surpasses several substantially \uline{larger open-source LLMs}, including Gemma2-27B-it, with gains of up to +8.1 on MATH and +0.6 on GSM8K, demonstrating that modular specialization can outperform sheer model scaling.
Compared with \uline{multi-teacher distillation} methods that rely on external supervision, \ourapproach—trained solely on self-generated data—achieves superior results in three out of four domains. In addition, it consistently outperforms \uline{model merging}, \uline{model grafting}, and \uline{routing-based MoE} baselines, including Twin-Merging, LoRA-MoE, and the prior self-alignment method \uline{Self-MoE}, with margins ranging from 1.2 to 10 points. These improvements stem from SkillWeaving’s core design: decomposing model capacity into domain-specialized skillpacks and reintegrating them through task-aware merging, which mitigates cross-task interference while maintaining low inference overhead.

\subsection{Alleviating interference and forgetting.}
Using the same training set, we compare against: \uline{Mixed-domain multi-task learning} and \uline{Continual multi-task learning} (trained sequentially from Math to Reasoning). We reveals that task-specific parameter updates exhibit low cosine similarity ($0.29$) and negative sign consistency ($-0.15$), indicating substantial conflict between domains. As a result, both mixed and sequential full fine-tuning inevitably suffer from task interference and catastrophic forgetting. In contrast, \ourapproach addresses this issue by explicitly separating domain-general and domain-specific knowledge, thus maintains consistent gains. Our merged backbone remains highly similar to all task vectors (e.g., cosine=0.153), suggesting that model merging effectively extracts shared capabilities.

\subsection{LLM-as-an-Agent Results}

We further evaluate \ourapproach in the LLM-as-Agent scenario, with results summarized in Appendix~\ref{app:llm_as_agent_result}.  
This setting is particularly well-suited to our modular design: 
a shared backbone remains active while capability-specific skillpacks are dynamically invoked depending on the user's tool calling.
We compare against two representative deployment strategies:
\vspace{-10pt}
\begin{enumerate}[leftmargin=2em]
    \item \textbf{Task-Specialized LLMs}: Deploying five separate 7B models for five tasks (totaling 5×7B parameters).
    \vspace{-10pt}
    \item \textbf{A Monolithic Model}: Using a single 32B model capable of handling all tasks jointly.
\end{enumerate}
\vspace{-5pt}
In contrast, \ourapproach requires only a single 7B backbone and five 0.5B skillpacks (totaling 9.5B parameters resident in memory), achieving significant savings in memory footprint. 
Crucially, the consistent use of a single 7B backbone and S-LoRA implementation for skillpacks allows for effective batch processing. 
As a result, \ourapproach delivers: 
\uline{1) \textbf{4.2× \emph{speedup}}} inference than the 32B monolith. \uline{2) \textbf{5.5× \emph{speedup}}} inference than the 5×7B model deployment. 
Despite this compact size and significant inference efficiency, \ourapproach achieves comparable performance: within 3\% of the task-specialized (5×7B) system, and within 5\% of the 32B monolithic model. 

These results confirm that \ourapproach is highly suited for agent-based applications, combining modularity, efficiency, and competitive task performance within a unified framework.

\section{Additional Results}

\subsection{Single-Skill Results as Ablation}
As shown in Tab.~\ref{tab:main_result}, we conduct ablation experiments across two dimensions: the choice of parameter format and the self-improvement algorithm. For each ablation, we replace the corresponding module in our pipeline with a competing alternative, in order to isolate and evaluate the effectiveness of each component in \ourapproach.



Our method consistently outperform \uline{PEFT}-based methods. These approaches are constrained by their limited trainable parameter space, suggesting that  \emph{full-parameter adaptation remains essential} for achieving strong task specialization.
Compared to \uline{ASVD}, our approach are observed a $30\%$ advantage in all domains. This further validates \emph{our full-tuning-then-zip design} of performing full-parameter fine-tuning followed by aggressive quantization, rather than relying solely on delta-based adapters.
And, compared to \uline{DeltaCome}, a recent delta-compression method, our approach achieves superior performance on 5 out of 8 tasks.

\vspace{-2pt}
Additionally, \ourapproach consistently outperforms \uline{Self-Specialization} and \uline{Self-Rewarding} across all tasks (e.g., +11 on Math and +7.9 on BBH). This gap largely stems from the limited reliability of self-judgments in prior methods, whereas \ourapproach selectively filters synthetic data and reinforces useful behaviors through DPO.



\subsection{\quantapproach Results}

\begin{figure}[h]
    \vspace{-10pt}
    \centering \includegraphics[width=1.0\linewidth, trim=10 150 60 0, clip]{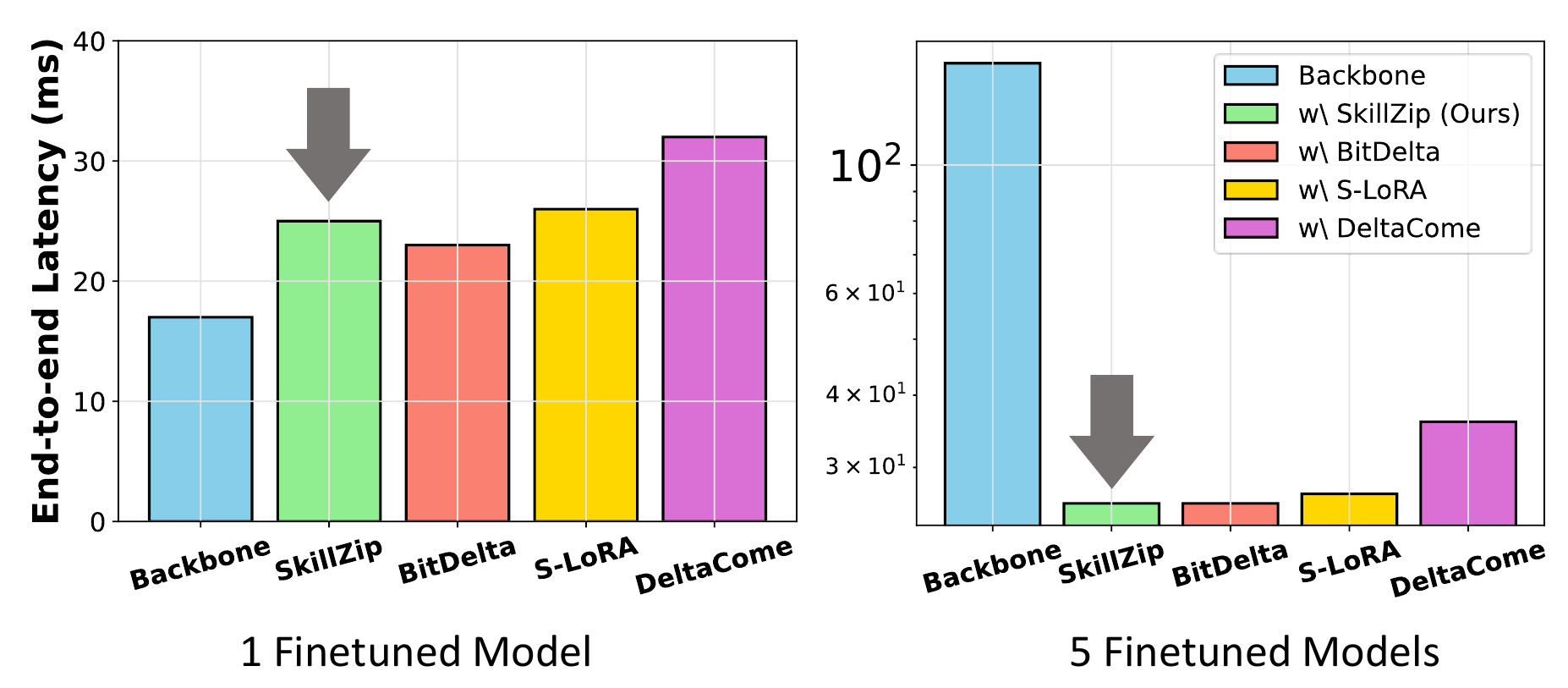}
    \vspace{-15pt}
    \caption{
    End-to-end inference latency comparison of delta-compression methods on a single A100-80G GPU using Qwen2.5-7B and its finetuned variants. The right plot shows the average token generation time for a single finetuned model, while the left plot reports the latency when serving five finetuned models concurrently. End-to-end latency includes full decoding time per token. And ``backbone'' denotes directly deploying multiple finetuned models simultaneously.
    \label{fig:latency}
    }
    \vspace{-10pt}
\end{figure}

\paragraph{Settings.}
We evaluate our quantization strategy under various configurations, each denoted as $\text{X}_{k_1}\text{A}_{k_2}\text{B}_{k_3}$. For instance, $X_8A_4B_4$ denotes using 8-bit activation $X$, 4-bit $A$, and 4-bit $B$. The \quantapproach configurations include $X_8A_8B_8$, $X_8A_4B_4$, $X_4A_4B_8$, and $X_4A_4B_4$.
We compare our method against prior compression baselines: BitDelta (denoted as $X_{16}W_1$), ASVD($X_{16}A_{16}B_{16}$), and DeltaCome. All experiments are conducted using Llama3.1-8B as the base model, with three model separately finetuned on three alignment tasks. Full results are presented in Tab.~\ref{tab:quant_result} and Figure~\ref{fig:latency}.

\begin{table}[ht] 
    \caption{Ablation study on \quantapproach and its comparison with other delta-compression baselines. 
    We evaluate compression fidelity across multiple settings and methods. 
    \label{tab:quant_result}
    }
    \vspace{-1mm}

  \centering
  \resizebox{1.0\linewidth}{!}{
 \begin{NiceTabular}{r|ccc|ccc|c}
      \toprule
      \rowcolor{gray!20} 
    \textbf{Setting} & \makebox[0.043\textwidth][c]{\textbf{Merge}}  & \makebox[0.05\textwidth][c]{\textbf{Smooth}} & \makebox[0.05\textwidth][c]{\textbf{Rotate}} & \makebox[0.045\textwidth][c]{MATH} & \makebox[0.045\textwidth][c]{MBPP} & \makebox[0.045\textwidth][c]{GPQA}  & \makebox[0.070\textwidth][c]{Average$\uparrow$ }
      \\   
      \midrule
      {Base}  & - & - & - & 51.9 & 66.9 & 33.6 & \cellcolor{dark1}50.8     \\
      {Target}  & - & - & - & 67.8 & 73.9 & 38.4 & \cellcolor{dark1}60.0    \\
      \midrule
      {$X_8A_8B_8$}  & \checkmark & \checkmark & \checkmark & \textbf{67.6} & \textbf{73.8} & \textbf{38.3} & \cellcolor{dark1}\textbf{59.8}    \\
      {$X_8A_8B_8$}   & \checkmark & \checkmark & \ding{55} & 66.6 & 73.6 & 38.0 & \cellcolor{dark1}59.4     \\
      {$X_8A_8B_8$}   & \checkmark & \ding{55} & \ding{55} & 63.0 & 72.9 & 37.2 & \cellcolor{dark1}57.7    \\
      {$X_8A_8B_8$}   & \ding{55} & \checkmark & \checkmark & 67.1 & 73.6 & 38.1 & \cellcolor{dark1}59.6     \\
    \midrule
    {$X_8A_4B_4$}  & \checkmark & \checkmark & \checkmark & 67.1 & 73.5 & 38.0 & \cellcolor{dark1}59.5     \\
    \midrule
    {$X_4A_4B_8$}  & \checkmark & \checkmark & \checkmark & 66.1 & 73.2 & 37.8 & \cellcolor{dark1}59.0     \\
    \midrule
    {$X_4A_4B_4$}  & \checkmark & \checkmark & \checkmark & 66.0 & 73.2 & 37.7 & \cellcolor{dark1}58.9     \\
    \midrule
    \multicolumn{4}{l}{DeltaCome\pub{NeurIPS24}} & 67.2 & 73.6 & 38.3 & \cellcolor{dark1}59.7     \\
    \midrule
    \multicolumn{4}{l}{$X_{16}A_{16}B_{16}$(SVD)}  & 61.0 & 71.1 & 36.2 & \cellcolor{dark1}56.3     \\
    \midrule
    \multicolumn{4}{l}{$X_{16}A_{8}B_{8}$(from DeltaCome)} & 63.0 & 72.1 & 37.0 & \cellcolor{dark1}57.4    \\
    \midrule
    \multicolumn{4}{l}{$X_{16}A_{4}B_{4}$(from DeltaCome)} & 65.4 & 73.2 & 37.6 & \cellcolor{dark1}58.7     \\
    \midrule
    \multicolumn{4}{l}{$X_{16}W_{1}$(BitDelta\pub{NeurIPS23})}  & 63.2 & 72.8 & 37.3 & \cellcolor{dark1}57.7    \\
    \midrule
    \multicolumn{4}{l}{$X_{16}A_{16}B_{16}$(ASVD)} & 623 & 72.6 & 37.3 & \cellcolor{dark1}57.0     \\
  \bottomrule
  \end{NiceTabular}
}
  \vspace{-5mm}
\end{table}
\paragraph{Ablation Study. }
We perform an in-depth ablation analysis on three technical components of our \quantapproach framework: \textit{1)} Model \emph{merg}ing for unified knowledge aggregation, \textit{2)} Channel-wise \emph{smooth}ing, \textit{3)} Rank-wise \emph{rotat}ion. 
We test combinations of these components and evaluate their contributions individually. Results in Tab.~\ref{tab:quant_result} show that each component contributes to performance gains, with the full combination yielding an average +2.1 improvement over the base quantization.

\paragraph{Fidelity and Latency.}
As summarized in Tab.~\ref{tab:quant_result}, our quantized skillpacks maintain comparable or better performance than previous approaches. Specifically, we achieve:

- A 2\% performance gain over DeltaCome,

- A 12\% improvement over BitDelta.

- A 15\% improvement over ASVD.

More importantly, our approach offers significant inference efficiency advantages. Figure~\ref{fig:latency} illustrates kernel-level latency (right) and end-to-end latency (left) across all methods. Thanks to our \emph{full quantization} of both weights and activations, we are able to directly compute in INT8 or INT4 formats using tensor core acceleration. This yields:

- 1.38× speedup over DeltaCome,

- 1.04× speedup over S-LoRA.

Although BitDelta uses even lower-bit quantization, its runtime dequantization and rehydration steps result in slower inference. Our approach slightly underperforms BitDelta in raw kernel latency (by a marginal 0.08×), but vastly outperforms it in overall task accuracy. This represents a superior trade-off between fidelity and latency, making our method both deployment-friendly and high-performing.



%



\section{Conclusion}
We present \ourapproach, a modular framework for improving LLMs under fixed memory and inference budget. By decomposing task capabilities into skill-specific modules, our approach enables scalable and interpretable improvement. Further, we introduce \quantapproach, a fully quantized delta-compression method that allows efficient deployment of specialized skills. Experiments across diverse benchmarks demonstrate that \ourapproach achieves superior performance and efficiency compared to existing self-training and model merging baselines. Overall, our results highlight modular skill composition as a promising direction for scalable and efficient enhancement of LLMs.

\section*{Acknowledgements}
This work was supported in part by National Natural Science Foundation of China (62476070), Shenzhen Science and Technology Program \seqsplit{(JCYJ20241202123503005, \, GXWD20231128103232001, \, ZDSYS20230626091203008, \, KQTD20240729102154066)}, Department of Science and Technology of Guangdong (2024A1515011540) and National Key R\&D Program of China (SQ2024YFE0200592).

\section*{Limitations}
\label{app:e}
While our proposed Skill Weaving framework demonstrates strong performance, parameter efficiency, and generality across diverse domains, we acknowledge several limitations that offer opportunities for further research.

First, our experiments primarily evaluate modular specialization on established benchmarks with clear domain boundaries. The effectiveness of SkillWeave in more open-ended or weakly defined task settings—such as emergent user behaviors or mixed-domain agentic workflows—requires further investigation.

Second, our improvement pipeline builds upon rule-based verification tailored for each domain, ensuring high-quality automatic feedback. While effective for structured tasks such as code generation or math reasoning, this verification paradigm becomes challenging in open-ended tasks—e.g., creative writing or abstract conversation—where no clear correctness criteria exist. 

Despite these limitations, our framework already lays a solid foundation for modular, scalable, and self-improving language models. We believe that by addressing the challenges of automatic skill discovery and verification in open-ended settings, Skill Weaving can be further evolved into a general-purpose capability engine adaptable to diverse real-world applications.

\section*{Ethical considerations}
This research adheres to established ethical standards in artificial intelligence and machine learning. All experiments were conducted using publicly available datasets or models under their respective licenses, and no personally identifiable or sensitive information was involved. The methods proposed are intended for academic and scientific purposes, with the goal of advancing understanding in machine learning rather than deployment in high-stakes decision-making without further safeguards.  
We recognize that advances in AI systems may pose potential societal risks, including issues of fairness, misuse, privacy, and environmental impact due to computational resource consumption. To mitigate these concerns, we emphasize responsible reporting of results, transparent acknowledgment of limitations, and a clear separation between research contributions and downstream applications.  
Future work building on this research should continue to assess possible ethical implications, particularly regarding bias, safety, and dual-use risks, and adopt appropriate measures to ensure beneficial and equitable outcomes.




\newpage

\appendix

\newpage
\appendix
~~~~~~~~~\textbf{{\Large Appendix for Skill Weaving}}
\vspace{0.7cm}

\section*{Reproducibility Statement}
Our implementation, including all code, training scripts, and evaluation datasets, is available at this anonymous-repo: \href{https://anonymous.4open.science/r/anonymous-repo-BFE7}{https://anonymous.4open.science/r/anonymous-repo-BFE7}

\section*{Overview}
This paper proposes a efficient self-improvement approach that decompose general-purpose LLMs into a collection of \textit{SkillPacks} that reorganize and refine the model’s internal capabilities. The appendix is structured according to our key contributions. 
We also make the project code available via an anonymous link for reproducibility: \href{https://anonymous.4open.science/r/anonymous-repo-BFE7}{https://anonymous.4open.science/r/anonymous-repo-BFE7}

\begin{itemize}
    \item Appendix~\ref{app:a} (Implementation Details) provides a detailed implementation description of rule-based verification and SKillZip design. 
    \item Appendix~\ref{app:b} (Additional Results) includes the results of agentic evaluation and results on smaller LLM.
    \item Appendix~\ref{app:c} (Related Works and Baselines) provides a detailed description of baseline and related works.
    \item Appendix~\ref{app:d} (Evaluation details) outlines the evaluation benchmarks and training datasets.
    \item Appendix~\ref{app:e} (Limitation and Future Work) provides a detailed dataset description.
\end{itemize}

\section{The Use of Large Language Models (LLMs)}
Throughout the preparation of this manuscript, large language models were employed exclusively for light stylistic refinement and the occasional grammatical adjustment. Every conceptual insight analytical thread, and interpretive conclusion emerged from the authors themselves; no algorithmic assistance was solicited for the framing, design, or substance of the work, and full scientific responsibility rests with the human contributors alone

\section{Implementation Details}
\label{app:a}
\subsection{Rule-based Verification}
To ensure the quality of self-generated data used for preference optimization, we design task-specific rule-based verification strategies tailored to each domain and dataset.
\paragraph{Mathematics.} For math-related datasets (e.g., GSM8K, MATH), we extract the final numerical answer from model outputs using regex-based pattern matching and compare it against the ground-truth solution. Only completions with exact matches are considered “helpful,” while mismatches are marked as “harmful.”
\paragraph{Code.}
In code generation tasks (e.g., HumanEval, MBPP), we execute the model-generated programs in a secure sandbox environment. A sample is accepted only if it passes all test cases and its output matches that of the reference solution. We also detect exceptions or infinite loops to identify invalid generations.
\paragraph{Reasoning.}
For tasks involving open-ended reasoning (e.g., ARC, BBH), we combine answer correctness with lightweight heuristic filters. These include rejecting overly short or excessively verbose answers, and favoring logically structured completions with sufficient diversity and specificity.
\paragraph{Dialogue.}
For instruction-following dialogue tasks (e.g., IFEval, AlpacaEval), we adopt two verification protocols for each type of prompts:

1) We design diverse prompting formats with explicit instruction constraints (e.g., “mention the keyword ‘AI’ at least 3 times”), and verify completions against corresponding structural rules—covering aspects such as keyword frequency, maximum length, repetition, banned tokens, format,  paragraph structure, language tone and so on. 
2) For loosely defined or open-ended instructions, we employ ensemble scoring from two widely adopted reward models to assess the quality of model completions. Each completion is evaluated across overall quality, and helpfulness, and harmlessness. We discard outliers whose average scores fall below the 10th percentile or exceed the 90th percentile to ensure balance and avoid preference bias. The reward models used are: 1) RewardModel-Mistral-7B-for-DPA-v1~\footnote{\url{https://huggingface.co/RLHFlow/RewardModel-Mistral-7B-for-DPA-v1}} 2) RLHFlow/ArmoRM-Llama3-8B-v0.1~\footnote{\url{https://huggingface.co/RLHFlow/ArmoRM-Llama3-8B-v0.1}}

\paragraph{Agent Tasks.}
For structured agent benchmarks such as \textbf{AgentBench} and \textbf{LifelongAgentBench}, we inherit and reuse the official rule-based evaluation criteria of each subdomain. For example, in the \textbf{Database} domain, a generation is marked correct only if the execution result of the generated SQL query matches the gold-standard output. Similar logic applies to \textbf{Operating System}, \textbf{Web}, and \textbf{Knowledge Graph} domains.

These rule-based verifications are fully automated and domain-specific, enabling efficient filtering of low-quality synthetic data prior to DPO training.

\subsection{SKillZip Implementation}

\paragraph{Hardware-Compatible Scaling Constraints.}
In hardware-accelerated GEMM~\citep{GEMM} kernels, FP16 scaling can only be efficiently applied along the outer dimensions—that is, before and after the INT8 matrix multiplication.
In our setting, this means the quantized multiplication: $\text{diag}(\vec{s}_A) \cdot \hat{A}  \cdot \hat{B} \cdot \text{diag}(\vec{s}_B)$, where $\hat{A}$ and $\hat{B}$ are INT8 or INT4 matrices, and $\vec{s}_A, \vec{s}_B$  are FP16 scaling vectors applied at the input and output channels, respectively.
Crucially, scaling between the two matrix multiplications (i.e., $\hat{A} \cdot \text{diag}(\vec{s}_A)  \cdot \text{diag}(\vec{s}_B)  \cdot \hat{B}$) is not supported by Tensor Core hardware and leads speed degradation.
Hence, we constrain quantization granularity as follows: per-token or per-tensor quantization for $\hat{X}$, per-tensor for $\hat{A}$, and per-tensor or per-channel for $\hat{B}$. The reconstructed output is computed by:
\begin{align}
Y = s_A \cdot \text{diag}(\vec{s}_X) \cdot \hat{X} \hat{A} \hat{B} \cdot \text{diag}(\vec{s}_B) .
\end{align}
where $s_A$ is a per-tensor scalar scaling factor and can be merged with $\vec{s}_X$  as  $\text{diag}(s_A\vec{s}_X)$.

\paragraph{Outlier-Induced Quantization Error.}
In the field of quantization, accuracy degradation often arises due to outliers — a small number of activation channels whose magnitudes are 100× larger than the rest. 
When misaligned with quantization axes, these outliers inflate the dynamic range, forcing most values in each quantization block to collapse to zero, thereby leading to significant quantization error. 

Consider the activation matrix $X\in \mathbb{R}^{T\times C_i}$, where each row corresponds to a token and each column corresponds to an input channel. In many LLMs activation, certain channels (say, the $i$-th column $X_{:,i}$) contain values with magnitudes exceeding 2000 across most or all tokens—this forms a channel-wise outlier. However, due to hardware constraints, activation quantization must be applied \emph{per-token}—i.e., along the row dimension, orthogonal to the direction in which outliers occur. This misalignment creates a quantization dilemma: each row vector $X_{j,:}$ acting as an independent quantization block, spans both extremely large outlier dimensions and low-magnitude, typical values (e.g., in the range $[-15, +15]$). When such a mixed-range vector is quantized uniformly into 8 bits, the outlier values dominate the scale and are mapped to the maximum bin (e.g., ±127), leaving the remaining values compressed into a narrow range near zero. As a result, most non-outlier values collapse to zero after quantization, effectively erasing useful information and causing severe precision loss.

A similar problem arises when quantizing low-rank weights $ W \approx AB$ 
Singular value decomposition (SVD) often concentrates most of the energy into the top few singular directions (i.e., the leading columns of $A$ and corresponding rows of $B$) forming rank-wise outliers, while the quantization is applied independently across matrix rows or columns.

\paragraph{Double Smoothing Strategy.}
To mitigate this, we propose a double smoothing strategy.

\textbf{\textit{1. Channel-wise Smoothing.}} We compute a domain-specific channel scaling vector $\vec{s} \in \mathbb{R}^{C_i}$ based on the average absolute activation magnitude per channel: $\vec{s}_i = f(\text{mean}(|X_{:,i}|))$, where $f(\cdot)$ is a monotonic mapping tuned for aggressive smoothing. We rescale:  $X \leftarrow X\cdot \text{diag}(s^{-1}), W\leftarrow \text{diag}(s) \cdot W$ thus transferring channel-wise quantization difficulty from $X$ to $W$.
Unlike SmoothQuant~\citep{smoothquant}, we apply more aggressive compression of outlier dimensions because: (1) the quantization burden is now shared across $X,A,B$ and each only bear one-third of the burden; and (2) each task domain $\tau_i$ has distinct activation statistics, for which we fit a unique scaling vector $\vec{s_i} = f(X^i)$

\textbf{\textit{2. Rank-wise Smoothing.}}
we apply truncated singular value decomposition (SVD) to the smoothed weight matrix $W_{\text{smooth}}$:
$W_{\text{smooth}} \approx U_R \Sigma_R V_R^T, \quad A = U_R \Sigma_R^{1/2}, \quad B = \Sigma_R^{1/2} V_R^T.$ And then apply a  second-stage rank-wise smoothing: a appropriate orthogonal rotation matrix $Q$ to disperses the energy evenly across dimensions:
$A_{\text{rot}} = A Q, \quad B_{\text{rot}} = Q^T B.$
The optimal $Q$ should satisfy:
\begin{itemize}
    \item $A Q$ and $Q^T B$ are individually easy to quantize;
    \item $X_\text{smooth}AQ$ is uniform, avoiding inner-product alignment between dominant rows of $X_\text{smooth}$ and columns of $A Q$, reducing precision loss when INT32 is truncated to INT8.
\end{itemize}
In practice, we find that randomly sampled orthogonal matrices suffice. We sample 10 candidates and select the one minimizing the final quantization loss.
 
Similar to DeltaCome~\citep{delta-come}, we also adopt GPTQ quantization~\citep{gptq} in the final step.

\paragraph{Hardware-Aware Inference.}
At inference time, we follow a fully quantized pipeline optimized for Tensor Core execution: 

1) Input $X$ is first smoothed using precomputed $s^{-1}$, and quantized to INT8/INT4 $\hat{X}_\text{smooth}$. 

2) We then load the quantized matrix , and perform the first INT8 GEMM: 
$\hat{X}_\text{smooth}\cdot\hat{A}_\text{smooth} \rightarrow \text{INT32}$. 
The output is intermediate result in INT32 format.

3) Rather than dequantizing the INT32 result, we aggressively truncate it to INT8—thereby preserving throughput and maintaining compatibility with the next GEMM. This truncated INT8 matrix is then used as input to a second matrix multiplication with $\hat{B}_\text{smooth}$, again using Tensor Core acceleration: $\text{INT8}\cdot\hat{B}_\text{smooth} \rightarrow \text{INT32}$. While, this second multiplication incorporates scaling vectors to recover the final FP16 output scale.

4) All scaling vectors and quantization parameters are precomputed to minimize runtime overhead.

5) To avoid latency from separate dequantization kernels, we fuse the dequantization step into the GEMM computation~\citep{bitblas}, utilizing fast-dequantization strategies~\citep{fast-dequant}.

This two-stage smoothing and hardware-aware execution pipeline forms the core of SkillZip, enabling low-latency, high-accuracy inference across diverse domains using low-rank, low-bit skillpacks.


\subsection{Routing Implementation}
\label{appendix:routing}
This section provides additional implementation details on how SkillWeave performs
domain routing during inference, how routing models are trained, and how routing
accuracy affects downstream performance. We additionally quantify the routing
overhead on inference speed. These clarifications address concerns regarding
domain identification when activating skillpacks at inference time.


SkillWeave adopts a routing mechanism fundamentally different from Mixture-of-Experts (MoE)
architectures: the backbone network is always activated, while a skillpack is activated
\emph{only} if the input belongs to its corresponding domain. As a result, SkillWeave does not
perform token-level expert selection but instead uses domain-level routing.

\paragraph{Agent Tasks.}
For agent-style tasks (e.g., tool selection, structured multi-step workflows), the domain
type of each request is inherently known in advance. Tool invocation APIs explicitly specify
which capability is being queried, and each skillpack corresponds to one such tool-specific or
capability-specific domain. Therefore, no domain classification is required, and routing is
deterministic.

\paragraph{General Tasks.}
For the general tasks evaluated in Table~\ref{tab:main_result}, we consider two settings:

\begin{enumerate}
    \item \textbf{Oracle Domain Labels.} For fair comparison with prior work, the main table
    assumes known domain labels, which is standard for multi-domain evaluation.

    \item \textbf{Learned Router.} To approximate realistic scenarios where domain labels may
    be unknown, we additionally train a domain classifier that predicts the skillpack to
    activate at inference time.
\end{enumerate}

\paragraph{Training the Routing Model}
We adopt Qwen2.5-0.5B as a lightweight routing model and attach a linear classification head.
The model is fine-tuned for sequence classification using 60{,}000 labeled prompts across
domains. Training is performed for 3 epochs on L40S GPUs, taking approximately one hour.

Routing accuracy is summarized in Table~\ref{tab:routing-accuracy}. The router achieves
exceptionally high performance across all domains, with accuracy, precision, and recall all
exceeding 0.999. Both false positive rate (FPR) and false negative rate (FNR) remain below
0.001. These results indicate that domain misclassification is exceedingly rare.

\begin{table*}[h]
\centering
\caption{Routing model performance across domains.}
\label{tab:routing-accuracy}
\begin{tabular}{lcccccc}
\toprule
Domain & Accuracy & FPR & FNR & Precision & Recall & F1 score \\
\midrule
Mathematics & 0.9987 & 0.0001 & 0.0013 & 0.9993 & 0.9987 & 0.9990  \\
Coding      & 1.0000 & 0.0000 & 0.0000 & 1.0000 & 1.0000 & 1.0000  \\
Dialogue    & 0.9990 & 0.0002 & 0.0010 & 0.9990 & 0.9990 & 0.9990  \\
Reasoning   & 1.0000 & 0.0001 & 0.0000 & 0.9992 & 1.0000 & 0.9996  \\
\bottomrule
\end{tabular}
\end{table*}

\paragraph{Impact of Routing on Downstream Task Performance}
We compare the downstream performance under two routing conditions:

\begin{itemize}
    \item \textbf{Oracle routing}: using ground-truth domain labels.
    \item \textbf{Model routing}: using the learned routing model.
\end{itemize}

Table~\ref{tab:routing-vs-oracle} shows that the performance difference between the two
settings is negligible across all benchmarks. This confirms that routing accuracy is
sufficiently high that it does not affect final task performance.

\begin{table*}[h]
\centering
\caption{Downstream performance with oracle vs.\ learned routing.}
\label{tab:routing-vs-oracle}
\resizebox{0.999\linewidth}{!}{
\begin{NiceTabular}{@{}l|cc|cc|cc|cc@{}}
    \toprule
        \rowcolor{gray!20}
        \multirow{2}{*}{\textbf{Method}} & \multicolumn{2}{c}{\textbf{Mathematics}} & \multicolumn{2}{c}{\textbf{Coding}} & \multicolumn{2}{c}{\textbf{Dialogue}} & \multicolumn{2}{c}{\textbf{Reasoning}}\\ 
        \cmidrule(lr){2-3} \cmidrule(lr){4-5} \cmidrule(lr){6-7} \cmidrule(lr){8-9}
        \rowcolor{gray!20}
        ~ & \textbf{GSM8k} & \textbf{MATH} & \makebox[0.08\textwidth][l]{\textbf{HumanEval}} & \makebox[0.08\textwidth][r]{\textbf{MBPP}} & \makebox[0.09\textwidth][l]{\textbf{AlpacaEval2}} & \makebox[0.08\textwidth][r]{\textbf{IFEval}} & \textbf{BBH} & \textbf{ARC-C} \\
        \midrule
\midrule
 \textbf{SKillWeave (Oracle Routing)} & 91.0 & 62.5 & 75.0 & 77.8 & 52.8 & 79.1 & 76.2 & 88.6 \\ \midrule
 \textbf{SKillWeave (Learned Routing)} & 91.0 & 62.3 & 75.0 & 77.8 & 52.9 & 78.9 & 76.2 & 88.6 \\
\bottomrule
\end{NiceTabular}
}
\end{table*}

\begin{figure*}[!h]
    \centering \includegraphics[width=0.7\linewidth, trim=20 0 0 50, clip]{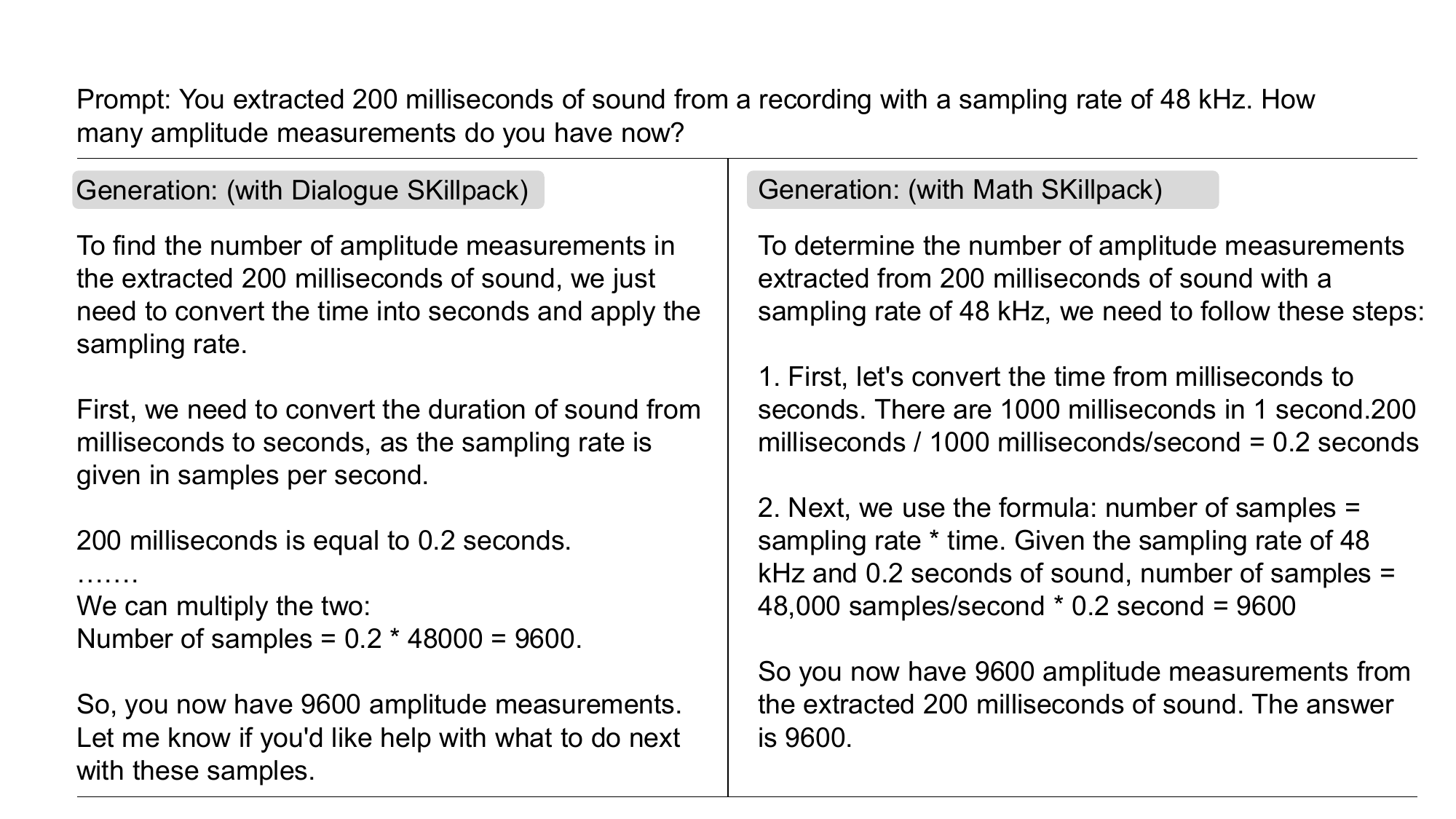}
    \vspace{0pt}
    \caption{
    An example of misidentification of learned router.
    \label{fig:routing-case}
    }
    \vspace{-10pt}
\end{figure*}

\paragraph{Case Study.}
We manually examined all samples where the router’s prediction differs from the oracle
domain. Interestingly, misclassified samples often contain mixed-domain content, such as a
dialogue prompt involving embedded mathematics. In such cases, the mathematical skillpack
can solve the underlying subproblem better than the dialogue model. Figure~\ref{fig:routing-case}
illustrates an example where routing to the “Math” skillpack yields higher-quality output
than routing to the “Dialogue” skillpack. This explains why routing mistakes do not degrade
overall accuracy.

\paragraph{Routing Overhead During Inference}
We further measure the inference-time overhead of routing. As reported in
Appendix~E.6, routing increases latency only marginally. The slowdown stems primarily from
\textbf{GPU memory contention} due to the routing model being resident in memory, rather than
from the routing model’s prefill or forward-pass computation.

Importantly, the routing model runs only once per request—not per token—so its cost is
effectively negligible compared to backbone decoding.



\section{Additional Results}
\label{app:b}
\begin{figure*}[h]
    \centering \includegraphics[width=1.0\linewidth, trim=0 1100 200 0, clip]{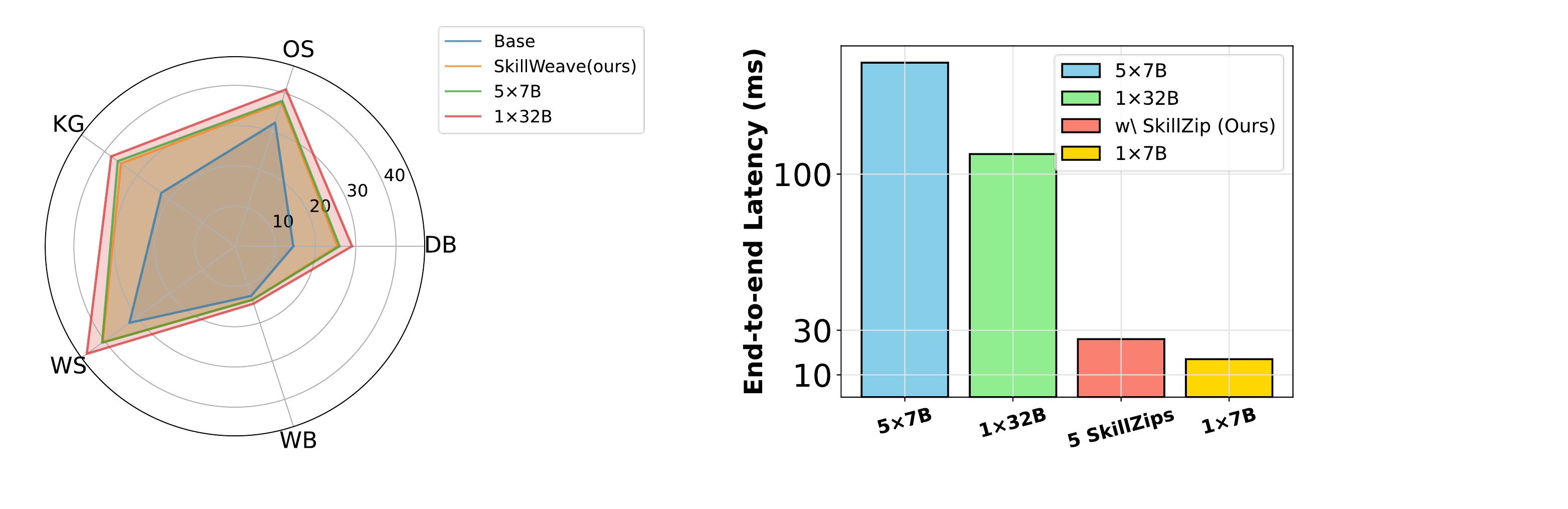}
    \vspace{-10pt}
    \caption{
    The left panel reports the performance across five dimensions of AgentBench under different configurations. The right panel compares the end-to-end inference latency when simulating agentic service calls, measured on a single A100-80G GPU. The 5×7B setting denotes five independently fine-tuned Qwen2.5-7B-Instruct models, each specialized for a different agent task, while 1×32B refers to a single monolithic Qwen2.5-32B-Instruct model.
    \label{fig:agent}
    }
    \vspace{-10pt}
\end{figure*}
\subsection{LLM-as-Agent Results}
\label{app:llm_as_agent_result}
Figure~\ref{fig:agent} illustrates the evaluation of SkillWeave in the LLM-as-Agent setting using Qwen2.5-7B-Instruct. The 5×7B configuration refers to five independently fine-tuned 7B models, each specialized for a distinct agent task, while 1×32B denotes a single monolithic Qwen2.5-32B-Instruct model. SkillWeave not only achieves significant inference acceleration, but also maintains competitive task performance, as detailed in the main text.

\begin{table*}[!t]
\caption{
Overall performance of SkillWeave in the general capability setting using Llama-3.2-1B-Instruct as the backbone. The best results for baselines are shown in bold and the second best are underlined.
\label{tab:app_result}
}
\centering
\resizebox{0.999\linewidth}{!}{
    \begin{NiceTabular}{@{}l|l|cc|cc|cc|cc@{}}
    \toprule
        \rowcolor{gray!20}
        \multirow{2}{*}{\textbf{Method}} & \multirow{2}{*}{\textbf{\#Params}} & \multicolumn{2}{c}{\textbf{Mathematics}} & \multicolumn{2}{c}{\textbf{Coding}} & \multicolumn{2}{c}{\textbf{Dialogue}} & \multicolumn{2}{c}{\textbf{Reasoning}}\\ 
        \cmidrule(lr){3-4} \cmidrule(lr){5-6} \cmidrule(lr){7-8} \cmidrule(lr){9-10}
        \rowcolor{gray!20}
        ~ & ~ & \textbf{GSM8k} & \textbf{MATH} & \textbf{HumanEval} & \textbf{MBPP} & \textbf{AlpacaEval2} & \textbf{IFEval} & \textbf{BBH} & \textbf{ARC-C} \\
        \midrule

        Llama3.2-1B-Instruct & 1.15B & 45.8	&32.3	&38.9	&49	&9.7	&58.4	&32.4	&69.4 \\   
        \midrule
        self-rewarding & 1.15B & 54.2	&32.6	&38	&44.6	&11.9	&63.2	&30.4	&71.1 \\
        \midrule
        Twin-merging	& 1.48B & 53.4	&33.8	&39.6	&49.4	&11.5	&61.9	&35.2	&70.1 \\
        \midrule
        FuseChat3.0	& 1.48B & \textbf{57.4}	&\textbf{36.2}	&\underline{41.4}	&\underline{45.5}	&\textbf{27.1}	&\textbf{72.5}	&\textbf{45.8}	&\textbf{77.5} \\
        \midrule
        \textbf{SKillWeave(Ours)}	& 1.42B & \underline{56.7}	&\underline{34.9}	&\textbf{42.5}	&\textbf{49.7}	&\underline{12.5}	&\underline{64.1}	&\underline{36.4}	&\underline{72.9} \\
        $\rightarrow$PEFT	& 1.42B & 46.1	&32.0	&39.1	&48.2	&12.1	&62.7	&34.8	&70.8 \\ \midrule
        
    \end{NiceTabular}
}
\vspace{-0.3cm}
\end{table*}

\subsection{Results on Other Model}
Table~\ref{tab:app_result} reports the general multi-capability evaluation results on \textsc{Llama3.2-1B-Instruct}, comparing SkillWeave against multiple baseline approaches. Under comparable model sizes, SkillWeave consistently outperforms all four baselines, clearly demonstrating its robustness and effectiveness in low-resource settings.

The self-rewarding approach performs poorly due to the limited evaluation ability of small models acting as judges, highlighting the necessity of our \emph{rule-based verification} framework for reliable self-improvement. Meanwhile, PEFT-based methods lag far behind—even on small models—due to their constrained trainable parameter space, further validating the superiority of our \emph{fully-finetune-then-compress (SkillZip)} strategy.

Although SkillWeave underperforms FuseChat3.0 in many domains, this is largely attributable to FuseChat3.0 leveraging stronger teacher models, which significantly outperform the \textsc{Llama3.2-1B-Instruct} student. 

\subsection{Inference Latency Measurement and Additional Results}
\label{appendix:latency}

This section provides detailed explanations of our end-to-end inference latency measurement protocol. 
We additionally include standardized latency tables reporting throughput
(tokens/s), normalized latency (ms/token), and kernel-level measurements across different batch sizes and sequence lengths.

\subsubsection{Experimental Setup}
\paragraph{Settings. }
All latency experiments are conducted on a single \textit{NVIDIA A100-80G} GPU.
As to inference engine, We adopt \textit{S-LoRA}\cite{s-lora} as our primary backend, because its implementation is most convenient for integrating our optimized low-rank kernels. Our methodology is compatible with vLLM and sGLang, which support similar execution models.
The base model is \textit{Llama3.1-8B-Instruct}.  
The testing Skillpack matrices have an average effective \textit{rank of 400} (ranging from 300 to 600 depending on the module).
As to request arrival process, 
we follow prior latency studies. The request stream is generated using a
\textit{Gamma arrival process} with coefficient of variation $=1$, which produces highly bursty arrival patterns representative of real-world workloads.
All measurements in Figure~3 and Figure~4 use this process.

\paragraph{Latency metrics.}
For each configuration we compute:
\begin{itemize}
    \item \textbf{Request throughput} (req/s)
    \item \textbf{Token throughput} (tokens/s)
    \item \textbf{Latency per generated token} (ms/token)
    \item \textbf{Time-to-First-Token} (TTFT)
    \item Distribution statistics: mean, median, P95, P99, min, max
\end{itemize}

\subsubsection{Baselines Details}
For the ``Backbone'' setting in Figure~3 and the ``5$\times$7B'' baseline in Figure~4,
we deploy multiple finetuned models using \textit{vLLM}, each as an
independent worker process under NVIDIA MPS.  
GPU memory is \emph{statically partitioned} among workers according to their
average request rate.

Because a single A100-80G cannot host more than five 7B--8B models simultaneously, we follow a realistic eviction-and-reloading policy: (1) track domain frequencies for past and projected future requests; (2) evict the model with the lowest combined usage score; (3) load the model with the highest expected usage into the freed memory.

We include this strategy for fairness: any alternative (e.g., sequential swapping) would only increase latency.  
Due to severe queuing delays and long tail delays, this strategy has already meant complete failure in reality, further highlighting the advantage of SkillWeave.

\subsubsection{End-to-End Latency Under Different Prompt and Generation Lengths}

We evaluate three groups of scenarios (full results in Table~6): \underline{(1)} Varying prompt length. \underline{(2)} Varying generation length. \underline{(3)} Varying request rate.

\begin{table*}[h]
\centering
\caption{End-to-end latency comparison.}
\label{tab:e2e-latency-details}

\resizebox{1.0\linewidth}{!}{
\begin{tabular}{ccc|ccc|cccccccc}
\toprule
\multicolumn{3}{c}{\textbf{Settings}} & \multicolumn{3}{c}{5×8B}  & \multicolumn{7}{c}{with 5 Skillpacks} \\ 
\midrule
request  & prompt & generation & request & token  & Latency  & request & token  & Latency & Latency & Latency & Latency &  Latency & Latency \\
rate  &  length &  length & throughput & throughput & (mean) & throughput & throughput & (mean) & (median) & (P95) & (P99) & (min) & (max) \\

\midrule
5	& 50	& 50	& 1.84 	& 101.98 	& 2.02  & 4.94 	& 274.18 	& 0.75 	& 0.75 	& 0.76 	& 0.76 	& 0.69 & 0.77 \\
5	& 200	& 50	& 1.43 	& 91.19 	& 2.62 & 4.94 	& 314.60 	& 0.76 	& 0.75 	& 0.76 	& 1.09 	& 0.69 & 1.10\\
5	& 500	& 50	& 1.39 	& 90.09 	& 2.70 & 4.94 	& 321.12 	& 0.76 	& 0.76 	& 0.76 	& 0.77 	& 0.69 & 0.77 \\
5	& 100	& 20	& 2.08 	& 51.02 	& 0.75 & 4.97 	& 121.86 	& 0.31 	& 0.31 	& 0.32 	& 0.32 	& 0.30 & 0.32\\
5	& 100	& 100	& 1.43 	& 159.23 	& 5.15 & 4.89 	& 543.92 	& 1.51 	& 1.51 	& 1.53 	& 1.53 	& 1.35 & 1.53\\
5	& 100	& 200	& 1.28 	& 307.48 	& 11.44 & 4.78 	& 1149.18 	& 3.06 	& 3.07 	& 3.09 	& 3.09 	& 2.72 & 3.10 \\
5	& 500	& 1000	& 0.46 	& 597.20 	& 103.84 & 2.32 	& 2983.21 	& 20.79 	& 21.33 	& 22.33 	& 22.40 	& 16.53 & 22.41 \\
5	& 500	& 4000	& 0.23 	& 920.68 	& 1857.59 & 1.42 	& 5691.31 	& 300.50 	& 300.49 	& 300.95 	& 301.00 	& 300.00 & 301.09 \\
0.5	& 500	& 4000	& 0.03 	& 122.13 	& 1411.42& 0.14 	& 572.80 	& 300.94 	& 300.94 	& 300.99 	& 301.09 	& 300.88 & 301.09 \\
1	& 500	& 4000	& 0.06 	& 238.64 	& 1441.09 & 0.29 	& 1142.88 	& 300.91 	& 300.91 	& 300.99 	& 300.99 	& 300.83 & 301.09 \\
3	& 500	& 4000	& 0.15 	& 612.56 	& 1677.45 & 0.85 	& 3419.07 	& 300.53 	& 300.55 	& 300.97 	& 300.99 	& 300.00 & 301.09 \\
5	& 500	& 4000	& 0.23 	& 920.68 	& 1857.59 & 1.42 	& 5691.31 	& 300.50 	& 300.49 	& 300.95 	& 301.00 	& 300.00 & 301.09 \\
7	& 500	& 4000	& 0.29 	& 1162.21 	& 2046.38 & 1.98 	& 7914.48 	& 300.50 	& 300.50 	& 300.95 	& 300.99 	& 300.00 & 301.09 \\
\bottomrule
\end{tabular}
}

\end{table*}

These experiments quantify end-to-end throughput and latency of SkillWeave and
naive baseline across a wide range of realistic workloads.

\subsubsection{Kernel-Level Latency: Prefill and Decode}

We further measure the \textbf{per-kernel latency} of our optimized low-rank
compute path. Using the Llama3.1-8B-Instruct \texttt{down\_proj}
matrix of shape $4096 \times 14336$, with rank~600, we benchmark:

\begin{itemize}
    \item the \textbf{prefill} phase with each sequence length $=1000$
    \item the \textbf{decode} phase with each sequence length $=1$
    \item batch sizes $\{1, 4, 10, 32\}$
\end{itemize}

Table~\ref{tab:kernel-latency} compares the standard \textbf{FP16 (X16W16)}
matrix multiplication kernel against our optimized \textbf{X8A8A8} kernel with
static quantization/dequantization. We also report performance under CUDA Graphs.

\begin{table*}[h]
\centering
\caption{Kernel latency comparison (ms). Left: FP16 X16W16. Right: Optimized X8A8A8.}
\label{tab:kernel-latency}
\begin{tabular}{c|ccc|ccc}
\toprule
\multirow{2}{*}{\textbf{Batch}} & \multicolumn{3}{c}{X16W16}  & \multicolumn{3}{c}{X8A8A8} \\ \cmidrule(lr){2-7}
 & \textbf{Prefill} & \textbf{Decode} & \textbf{Decode (Graph)} & \textbf{Prefill} & \textbf{Decode} & \textbf{Decode (Graph)} \\  
\midrule
1  & 0.6673 & 0.1167 & 0.1086 & 0.1698 & 0.0558 & 0.0271 \\
4  & 2.0216 & 0.1064 & 0.0945 & 0.4669 & 0.0525 & 0.0290 \\
10 & 4.9063 & 0.1064 & 0.0991 & 1.1168 & 0.0524 & 0.0291 \\
32 & --     & 0.1082 & 0.0963 & --     & 0.0553 & 0.0280 \\
\bottomrule
\end{tabular}
\end{table*}

The optimized kernel is significantly faster than the backbone’s FP16 path,
and the incremental latency introduced by activating a single skillpack is negligible.






\section{Related Works and Baselines}
\label{app:c}

This section provides detailed descriptions of the related works and baseline methods used in our study.
Some of these works also serve as baselines in our experiments and are marked with a superscript \textsc{asterisk (*)} next to their names.\footnote{Beyond the directly comparable methods reviewed in this section, our modular skillpack design is also inspired by broader lines of research on evolutionary parameter optimization for deep neural networks~\citep{du2024impacts}, structured perception decomposition in low-level vision~\citep{du2023end}, and prototype- and memory-based reasoning for multimodal vision–language understanding in specialized domains~\citep{zhu2025pathology,zhu2026medeyes,lin2026medcausalx,zhu2026medsynapsevbridgingvisualperception}.}


\subsection{Self-Improvement for LLM}
Self-improvement methods aim to enhance language models using only their own generations, without relying on external human annotations or teacher supervision. These approaches explore how LLMs can autonomously refine their capabilities via self-generated data and internal feedback mechanisms.
\begin{itemize}
\item \textbf{Self-Specialization*}~\citep{self-special} finetunes an LLM on its own task-specific generations to induce latent expertise. However, it does not distinguish between high- and low-quality outputs, resulting in unstable performance and potential error reinforcement.
\item \textbf{Self-rewarding*}~\citep{self-rewarding} introduces LLM-as-a-judge prompting, where the model scores its own responses using handcrafted criteria and then applies preference optimization (e.g., DPO) to favor high-quality generations.
\item \textbf{Meta-rewarding*}~\citep{meta-rewarding} further enhances the self-rewarding method by improve both acting and judging skills of models simultaneously.
\item \textbf{Self-Align*}~\citep{Self-Align} presents human-written alignment principles to the model, which are then internally triggered to filter and guide generation. It offers a lightweight and interpretable way to regulate self-improvement using task-level rules.
\item \textbf{Self-MoE*}~\citep{self-moe} 
extends Self-Specialization into a modular multi-task setting by training separate LoRA experts for each domain. These LoRA modules are then assembled into a LoRA-MoE model (in the LoRAHub style) to achieve compositional generalization across tasks.
\item \textbf{RLAIF}~\citep{RLAIF} 
reuses the RLHF pipeline but replaces reward signals with synthetic preferences generated by LLMs themselves. It provides a scalable way to apply reinforcement learning without manual annotations. 
\end{itemize}
We adopt \textbf{Self-Rewarding} and \textbf{Self-Align} as multi-skill baselines in our experiments. Following their original setups, we design distinct prompting templates and scoring criteria for each task domain. However, instead of training each domain in isolation, we merge all task datasets and perform joint training using both SFT and DPO objectives over the combined data—ensuring consistency and comparability with SkillWeave.

We also include \textbf{Self-Rewarding} and \textbf{Self-Specialization} as single-skill baselines, replacing the self-improvement module in our pipeline with these alternatives. For fairness, we retain our full fine-tuning and SkillZip compression stages, allowing us to isolate and evaluate the effectiveness of the self-improvement component itself. The corresponding results are reported in Table.1.

\subsection{Model Merging and Grafting}
Model Merging aims to combine multiple task-specific models into a single multitask model by algebraically manipulating their finetuned parameters. By scaling and summing the deltas from different tasks, merging methods seek to amplify beneficial knowledge while mitigating harmful interference across tasks. These approaches attempt to resolve parameter conflicts and redundancies to form a unified and robust multitask representation.
Model Grafting~\citep{grafting}, on the other hand, takes a more surgical approach. It selectively transplants a small subset of task-specific parameters into the pre-trained model, aiming to recover finetuned performance while introducing minimal overhead. This paradigm emphasizes the localization and reuse of transferable skills across tasks.
\begin{itemize}
\item \textbf{Task Arithmetic*}~\citep{ta} 
first introduces the concept of \emph{``task vectors''}—the difference between a finetuned model and its base—and proposes to merge them via linear operations: $\theta_\textrm{merge} = \theta_\textrm{init} + \lambda * \sum_{t=1}^n \tau_t$, where $\tau_t$ is the task vector for task $t$.
\item \textbf{AdaMerging}~\citep{adamerging} 
extends task arithmetic by automatically learning optimal linear merging coefficients to adaptively tune layer-wise or task-wise weights. 
It uses entropy minimization on unlabeled evaluation data as a surrogate objective, enabling unsupervised merging.
\item \textbf{Ties-Merging*}~\citep{ties} further solves the task conflict problem in Task Arithmetic~\citep{ta} by explicitly resolving parameter conflicts via a three-stage process: Trim redundant parameters, Elect,  and Disjoint Merge to isolate interference.
\item \textbf{PCB-Merging*}~\citep{pcb} effectively adjusts parameter coefficients through balancing parameter competition within model population.
\item \textbf{FR-Merging}~\citep{free-merge} 
emphasizes the importance of merging common capabilities into the backbone before combining task-specific skills, thereby preserving general knowledge.
\item \textbf{DARE*}~\citep{dare} sets the majority of delta parameters to zero and rescale the rest by \( \theta' = \theta \cdot (1/(1-p)) \) where \( p \) is the proportion of delta parameters dropped, therefore efficiently reduces parameter redundancy.
\item \textbf{TALL-MASK*}~\citep{talls} localize the task-specific information in a multi-task vector, which deactivates irrelevant parts for each task in the merged multi-task vector with binary masks.
\item \textbf{EMR-Merging*}~\citep{emr} first selects a unified model from all weights, then generates lightweight task-specific modulators—masks and rescalers—to align direction and magnitude with each source model.
\item \textbf{Model Grafting}~\citep{grafting} 
introduces the notion of skill localization by identifying which parameter subsets are critical for each task. It then selectively grafts these modules onto the pre-trained model, achieving task recovery without full model duplication.
\item \textbf{Knowledge Fusion by Evolving Weights}~\citep{du2024knowledge} formulates the integration of multiple finetuned language models as an evolutionary search over weight populations, combining task-specific knowledge through evolutionary operators without relying on external supervision.
\item \textbf{GraftLLM}~\citep{add2graftllm} couples modular skillpacks with knowledge fusion to integrate diverse expert capabilities into a single LLM while preserving per-skill expertise, providing a precursor view of the modular decomposition pursued in this work.
\item \textbf{Neural Parameter Search}~\citep{add1NeuralParameter} learns task-specific parameter masks via differentiable search to produce slimmer fine-tuned models with stronger cross-task transfer, complementing merging-based compression strategies.
\item \textbf{Dynamic Model Merging}~\citep{du2026didi} adapts merging coefficients at inference time based on the input distribution, enabling slim multi-skill serving without retraining the merged model for each deployment.
\item \textbf{Neuron-aligned Merging}~\citep{fang2025disentangling} disentangles task interference at the neuron level and aligns merging operations with neuronal activation mechanisms, mitigating conflicts that pure weight-space arithmetic cannot resolve.
\item \textbf{MMER}~\citep{mmer} extends parameter merging to multi-modality expansion by decoupling modality-specific deltas from a shared backbone, demonstrating that the merging-and-decoupling pattern generalizes beyond unimodal task vectors.
\end{itemize}
In our evaluation, all model merging and grafting baselines are instantiated using the same pre-trained model and identical task-specific finetuned models produced by SkillWeaving. To ensure fair comparison, we report that TALL-MASK and EMR-Merging introduce significantly more parameters than SkillWeave due to the inclusion of large task-specific components. 

\subsection{LoRA-based MoE}
LoRA-based Mixture-of-Experts (MoE) models combine the modularity and specialization advantages of MoE architectures with the parameter efficiency of Low-Rank Adaptation (LoRA). In these models, each expert is implemented as a lightweight LoRA adapter, typically applied to all major modules within every Transformer block. This design enables fine-grained task specialization while significantly reducing memory and training overhead compared to traditional dense experts.
\begin{itemize}
\item \textbf{LoRA-MoE*}~\citep{LoraMoE1} 
 extends the standard MoE framework by deploying multiple LoRA experts in parallel across a multi-task training setup. It introduces an expert-balancing mechanism to ensure that all LoRA modules are utilized effectively. 
\item \textbf{LoRAHub}~\citep{LoraMoE1} employs Low-rank Adaptations to dynamically combine task-specific modules for cross-task generalization, and adapts to new tasks by configuring \( \theta' = \sum_{k=1}^{K} w_k \cdot \theta_k \).
\item \textbf{Twin-Merging*}~\citep{twin} compress multiple finetuned models into a compact LoRA-MoE format by merging, singular value decomposition and pruning. Then a trainable router is used to dynamically select among them. 
\item \textbf{Hierarchical-Expert Alignment}~\citep{li2025multi} aligns LLMs against multiple objectives by routing among a hierarchy of LoRA experts, each specialized for a different reward dimension, illustrating how modular expert structures benefit multi-objective preference optimization.
\end{itemize}

\subsection{Delta Compression for LLM}
Delta Compression aims to to reduce the overhead of storing and serving multiple task vectors.
Delta compression explores the idea that fine-tuning introduces sparse and structured modifications to a pre-trained model, which can be efficiently compressed. These compressed deltas, when combined with a shared base model, enable storage and inference efficiency by avoiding full model duplication.
\begin{itemize}
\item \textbf{BitDelta*}.~\citep{bitdelta} proposes a post-training method that directly quantizes finetuned deltas to 1-bit precision. This strategy reduces both storage and latency while maintaining acceptable fidelity. 
\item \textbf{DeltaCome*}.~\citep{delta-come} extends delta compression by first applying singular value decomposition (SVD) to each delta and then employing varying bit-widths quantization for different singular vectors based on their singular values. 
\item \textbf{ASVD*}.~\citep{asvd} introduces Activation-aware SVD, a training-free SVD method that improves decomposition accuracy by transforming weight matrices using activation outlier statistics. 
Although originally proposed for backbone compression, we adapt ASVD to delta compression in our evaluation. 
\item \textbf{D-QRELO}~\citep{li2026d} provides training-free and data-free delta compression by combining quantization with residual low-rank approximation, jointly reducing the storage and serving cost of multiple finetuned LLMs without access to calibration data.
\end{itemize}
In our experiments, all delta compression baselines are built using the same base model and target finetuned models through SkillWeave, ensuring a controlled comparison. 

\subsection{Quantization for LLM}
Quantization aims to reduce the bitwidth of model parameters and activations to improve inference speed and reduce memory usage. While weight-only quantization offers moderate savings, full quantization—compressing both weights and activations—is essential to eliminate runtime decompression and unlock true speedups on hardware accelerators.
One major challenge in quantization is handling outliers—activation  values that are orders of magnitude larger than the rest. These outliers distort the dynamic range and lead to large quantization errors, especially when misaligned with quantization axes. Recent research has focused on outlier-aware quantization to preserve accuracy while enabling aggressive bit reduction.
\begin{itemize}
\item \textbf{GPTQ}~\citep{gptq} 
is a post-training quantization method that minimizes quantization error by greedily adjusting non-quantized parameters. It is particularly well-suited for LLMs and supports weight-only quantization.
\item \textbf{GPTZip}~\citep{gptzip} 
extends GPTQ to finetuned deltas, allowing the same quantization process to compress skill-specific updates.
\item \textbf{LLM.int8}~\citep{llmint8} introduces an 8-bit weight-only quantization framework that use vector-wise quantization to quantize most of the features and isolates the outlier feature dimensions into a 16-bit matrix multiplication for the emergent outliers.
\item \textbf{AWQ}~\citep{awq} focuses on identifying and protecting salient weight channels using activation outlier statistics. By scaling these channels pre-quantization, AWQ avoids expensive mixed-precision inference and retains performance with pure INT8 computation. 
\item \textbf{SmoothQuant}~\citep{smoothquant} proposes a full INT8 quantization (W8A8) technique that migrates quantization difficulty from activations to weights via mathematically equivalent transformations. 
\end{itemize}

\subsection{Multi-Teacher Distillation}
Multi-Teacher Distillation extends classical model distillation by allowing a student model to simultaneously learn from multiple teacher models. Rather than assuming any single teacher is universally superior, this approach aims to distill specialized capabilities from each teacher—leveraging their complementary strengths to form a more holistic student.
Recent methods in this line of work typically combine supervised fine-tuning (SFT) with preference-based learning objectives such as Direct Preference Optimization (DPO) or WRPO~\citep{wrpo}. This hybrid strategy enables the student model to mimic helpful teacher behaviors while suppressing harmful self-generations, promoting both alignment and robustness.
\begin{itemize}
\item \textbf{FuseLLM*}~\citep{wan2024knowledge} is the first to introduce multi-teacher distillation for fusing knowledge from heterogeneous large language models of different scales and structures.
\item \textbf{FuseChat2.0}~\citep{fusechat2} 
refines this idea by a statistics-based token alignment for compatibility. It uses lightweight pairwise fine-tuning into target models of the same size and merges the targets in parameter space.
\item \textbf{FuseChat3.0*}~\citep{fusechat3} further introduces implicit model fusion and a DPO-based strategy to enhance alignment and integration performance across heterogeneous LLMs.
\end{itemize}
We faithfully reproduce these baselines using the official FuseChat-3.0 implementation available at \href{https://github.com/SLIT-AI/FuseChat-3.0}{SLIT-AI/FuseChat-3.0}. Although our experimental domains differ from those in the original works, we strictly reuse the same teacher models and student architecture to ensure maximum fidelity in reproduction. 

\section{Experiment Details}
\label{app:d}
\subsection{Evaluation Benchmarks}
\textbf{AlpacaEval-2}~\citep{AlpacaEval} evaluates instruction-following ability using 805 prompts from five datasets, measured by raw win rate (WR)~\citep{dubois2024length} and length-controlled win rate (LC). GPT-4-Preview-1106 serves as both the reference and judge; we report WR in main text and LC in Appendix.

\textbf{IFEval}~\citep{IFEval} \emph{(Strict, O shot, CoT) }assesses LLMs with automatically verifiable instructions, such as length constraints or required keywords. Evaluation is performed via rule-based parsing, enabling scalable and objective instruction-following assessment.

\textbf{GSM8K}~\citep{cobbe2021gsm8k}\emph{(O shot, CoT) } is a set of grade-school math word questions that evaluates mathematical reasoning capabilities.

\textbf{MATH}~\citep{hendrycks2021math}\emph{(O shot, CoT) } is a dataset of math problems ranging in difficulty from middle school to high school competition level. It tests a wide range of mathematical skills, including algebra, calculus, number theory, and probability.

\textbf{HumanEval}~\citep{chen2021evaluating}\emph{(O shot, CoT) } evaluates code generation capabilities by presenting models with function signatures and docstrings and requiring them to implement the function body in Python.

\textbf{MBPP}~\citep{austin2021program}\emph{(O shot, CoT) } is a dataset of simple programming problems designed to assess the ability of models to generate short Python code snippets from natural language descriptions.

\textbf{BBH}~\citep{bbh}\emph{(1 shot, CoT) } (Big Bench Hard) targets multi-step logical reasoning and compositional generalization through 23 hand-crafted tasks under few-shot settings.

\textbf{ARC-C}~\citep{arc}\emph{(O shot, CoT) } contains challenging science multiple-choice questions filtered to exclude retrieval or co-occurrence-based solutions, promoting higher-order QA reasoning.

\textbf{AgentBench}~\citep{agentbench} evaluates agentic reasoning across diverse interactive tasks in executable environments. Each task measures success rate or step success rate under ReAct-style prompting.

\subsection{Training Datasets}
We construct our training set from diverse sources covering a broad spectrum of skills, domains, and instruction styles. The dataset includes both human-written and model-generated examples and overlaps significantly with \textsc{FuseChat-Mixture}~\citep{fusechat2}.

The full list of training data sources and construction methods is as follows: 
\begin{itemize}
    \item \textbf{Math}: 
    OpenMathInstruct-2~\footnote{\url{https://huggingface.co/datasets/nvidia/OpenMathInstruct-2}}, MetaMathQA~\footnote{\url{https://huggingface.co/datasets/meta-math/MetaMathQA}}, AMC 23~\footnote{\url{https://huggingface.co/datasets/AI-MO/aimo-validation-amc}}
    \item \textbf{Code}: Self-Oss-Instruct-SC2~\footnote{\url{https://huggingface.co/datasets/bigcode/self-oss-instruct-sc2-exec-filter-50k}}, OSS-Instruct~\footnote{\url{https://huggingface.co/datasets/ise-uiuc/Magicoder-OSS-Instruct-75K}}, Evol-Alpaca~\footnote{\url{https://huggingface.co/datasets/theblackcat102/evol-codealpaca-v1}}, Python-Code~\footnote{\url{https://huggingface.co/datasets/ajibawa-2023/Python-Code-23k-ShareGPT}}.
    \item \textbf{Dialogue}: 
    Magpie-Pro-DPO~\footnote{\url{https://huggingface.co/datasets/Magpie-Align/Magpie-Llama-3.1-Pro-DPO-100K-v0.1}},
    \textbf{Orca-Best}\footnote{\url{https://huggingface.co/datasets/shahules786/orca-best}}, \textbf{Capybara}\footnote{\url{https://huggingface.co/datasets/LDJnr/Capybara}}, UltraFeedback~\footnote{\url{https://huggingface.co/datasets/princeton-nlp/llama3-ultrafeedback-armorm}}, HelpSteer2~\footnote{\url{https://huggingface.co/datasets/nvidia/HelpSteer2}}, HelpSteer~\footnote{\url{https://huggingface.co/datasets/nvidia/HelpSteer}}, ShareGPT-GPT4~\footnote{\url{https://huggingface.co/datasets/shibing624/sharegpt_gpt4}}.
    \item \textbf{Reasoning}: ARC~\footnote{\url{https://huggingface.co/datasets/allenai/ai2_arc}}, BBH~\footnote{\url{https://github.com/suzgunmirac/BIG-Bench-Hard}}.
    \item \textbf{Agentic}: AgentBench~\footnote{\url{https://github.com/THUDM/AgentBench}},
Mind2Web~\footnote{\url{https://github.com/OSU-NLP-Group/Mind2Web}},
WebShop~\footnote{\url{https://github.com/princeton-nlp/webshop}},
AgentInstruct~\citep{AgentInstruct}, AgentBoard~\footnote{\url{https://github.com/hkust-nlp/AgentBoard}}.
\end{itemize}

\subsection{Hyperparameter Settings}
 
\begin{table*}[h]
\centering
\caption{Hyperparameters for various tasks on Llama-3.1-8B-Instruct model during online DPO stages.}
\label{tab:trainig_hyperparameters}
\resizebox{0.7\linewidth}{!}{
    \begin{tabular}{lcccc}
    \toprule
    \textbf{Target Task} & \textbf{Epochs} & \textbf{DPO Learning Rate} & \textbf{DPO $beta$} & \textbf{DPO Loss Type} \\
    \midrule
    Math & 3 & $1 \times 10^{-6}$ & $10\sim12$ & $\mathcal{L}_{\text{LN-DPO}}$ \\
    Coding & 5 & $5 \times 10^{-7}$ & $10\sim12$ & $\mathcal{L}_{\text{LN-DPO}}$ \\
    Dialogue & 5 & $1 \times 10^{-6}$ & $5\sim8$ & $\mathcal{L}_{\text{LN-DPO}}$ \\
    Reasoning & 3 & $9 \times 10^{-7}$ & $8$ & $\mathcal{L}_{\text{LN-DPO}}$ \\
    \bottomrule
    \end{tabular}
}
\end{table*}

In DPO experiments, we utilize the TRL library\footnote{\url{https://github.com/huggingface/trl}} as the training framework for online DPO. We train the LLMs using a batch size of 128 and a maximum length of 4096 on a single node with 8x80GB NVIDIA A800 GPUs. wW use AdamW optimizer with $\beta_{1}=0.9$ and $\beta_{2}=0.999$, weight decay=0.1 with cosine decay and warmup ratio 0.03, BF16 mixed precision, gradient-norm clip 1.0. We truncate sequences to a context length 4096 tokens (prompt+response); padding is left-aligned.

For online sampling (training data generation) we use vLLM with temperature=0.7, top-0.95
p=0.95, top-k=50, n=64 candidates per prompt, max new tokens=2048, and a mild repetition penalty=1.05. At test time, we use greedy decoding (temperature=0)

Because we train separately per task, we select task-specific hyperparameters (\eg learning rate, epochs, $\beta$. The hyperparameter configurations for different tasks are detailed in Tab.~\ref{tab:trainig_hyperparameters}. In one word, we adopt a simple yet robust tuning strategy as a general rule: 
\begin{itemize}
    \item For tasks prone to destabilizing training, we reduce the learning rate.
    \item For inherently harder tasks, we increase the number of online rounds. 
    \item For task where synthetic samples exhibit large distributional variance, we employ a higher $\beta$. 
\end{itemize}

Because all preference pairs are self-generated, the reference term $\log \pi_{ref}(y|x)$ has limited regularization value; we therefore adopt a LN-style~\citep{simpo} objective (i.e., DPO with length normalization and without an explicit reference log-ratio)







\end{document}